%% file: main.tex
\documentclass[DIV15, a4paper]{scrartcl}

\usepackage[ruled, linesnumbered, vlined, commentsnumbered]{algorithm2e}
\usepackage{authblk}
\usepackage{prettyref}
\usepackage{graphicx}
\usepackage{subfig}
\usepackage{booktabs}
\usepackage{colortbl}
\usepackage{amssymb}
\usepackage{amsfonts}
\usepackage{amsmath}
\usepackage{amsthm}
\usepackage{multirow}
\usepackage{tabularx}
\usepackage{footnote}
\usepackage{threeparttable}
\usepackage{hyperref}
\usepackage{amsopn}
\usepackage{cite}
\usepackage{subfig}
\usepackage{setspace}
\usepackage{color}
\usepackage{hhline}

\usepackage{tikz}
\usepackage{pgfplots}
\usetikzlibrary{positioning,shapes,shadows,arrows,calc}
\tikzstyle{component}=[rectangle, draw=black, rounded corners, fill=blue!40, drop shadow, text centered, anchor=north, text=white, minimum height=1cm]
\tikzstyle{arrow}=[->, thick]

\pgfplotsset{compat=1.12}
\usetikzlibrary{intersections}
\usetikzlibrary{pgfplots.statistics}
\usepgfplotslibrary{fillbetween}

\usepackage{xcolor}  
\definecolor{myblue}{RGB}{34,31,217}

\definecolor{mycyan}{gray}{.7}

\DeclareMathOperator*{\argmax}{argmax}
\DeclareMathOperator*{\argmin}{argmin}

\newcommand{\bb}[1]{\multicolumn{1}{>{\columncolor{mycyan}}c}{\textbf{{#1}}}}

\newcommand{\pref}{\prettyref}

\newrefformat{fig}{Fig.~\ref{#1}}
\newrefformat{tab}{Table~\ref{#1}}
\newrefformat{sec}{Section~\ref{#1}}
\newrefformat{app}{Appendix~\ref{#1}}
\newrefformat{alg}{Algorithm~\ref{#1}}
\newrefformat{property}{Property~\ref{#1}}
\newrefformat{theorem}{Theorem~\ref{#1}}
\newrefformat{corollary}{Corollary~\ref{#1}}
\newrefformat{proposition}{Proposition~\ref{#1}}
\newrefformat{def}{Definition~\ref{#1}}
\newrefformat{eq}{equation~(\ref{#1})}

\usepackage{graphicx}
\definecolor{Gray}{gray}{0.9}
\usepackage{scrpage2}

\usepackage{lscape}

\begin{document}

\title{\textbf\LARGE\fontfamily{cmss}\selectfont Two-Archive Evolutionary Algorithm for Constrained Multi-Objective Optimization\footnote{This paper is submitted for possible publication. Reviewers can use this manuscript as an alternative in peer review.}}

\author[1]{\normalsize\fontfamily{lmss}\selectfont Ke Li$^{\#}$}
\author[2]{\normalsize\fontfamily{lmss}\selectfont Renzhi Chen$^{\#}$}
\author[3]{\normalsize\fontfamily{lmss}\selectfont Guangtao Fu}
\author[2]{\normalsize\fontfamily{lmss}\selectfont Xin Yao}
\affil[1]{\normalsize\fontfamily{lmss}\selectfont Department of Computer Science, University of Exeter}
\affil[2]{\normalsize\fontfamily{lmss}\selectfont CERCIA, School of Computer Science, University of Birmingham}
\affil[3]{\normalsize\fontfamily{lmss}\selectfont Department of Engineering, University of Exeter}
\affil[$\ast$]{\normalsize\fontfamily{lmss}\selectfont Email: \{k.li, g.fu\}@exeter.ac.uk, \{rxc332, x.yao\}@cs.bham.ac.uk}
\affil[$\#$]{\normalsize\fontfamily{lmss}\selectfont The first two authors make equal contributions to this paper.}

\renewcommand\Authands{ and }

\date{}
\maketitle

{\normalsize\fontfamily{lmss}\selectfont\textbf{Abstract: } }When solving constrained multi-objective optimization problems, an important issue is how to balance convergence, diversity and feasibility simultaneously. To address this issue, this paper proposes a parameter-free constraint handling technique, a two-archive evolutionary algorithm, for constrained multi-objective optimization. It maintains two co-evolving archives simultaneously: one, denoted as the convergence archive, is the driving force to push the population toward the Pareto front; the other one, denoted as the diversity archive, mainly tends to maintain the population diversity. In particular, to complement the behavior of the convergence archive and provide as much diversified information as possible, the diversity archive aims at exploring areas under-exploited by the convergence archive including the infeasible regions. To leverage the complementary effects of both archives, we develop a restricted mating selection mechanism that adaptively chooses appropriate mating parents from them according to their evolution status. Comprehensive experiments on a series of benchmark problems and a real-world case study fully demonstrate the competitiveness of our proposed algorithm, in comparison to five state-of-the-art constrained evolutionary multi-objective optimizers.

{\normalsize\fontfamily{lmss}\selectfont\textbf{Keywords: } }Multi-objective optimization, constraint handling, evolutionary algorithm, two-archive strategy

\input{introduction}

\input{related}

\input{proposal}

\input{settings}

\input{experiments}

\input{conclusion}

\section*{Acknowledgment}
This work was supported by the Ministry of Science and Technology of China (Grant No. 2017YFC0804002), the Science and Technology Innovation Committee Foundation of Shenzhen (Grant No. ZDSYS201703031748284) and EPSRC (Grant No. EP/J017515/1).

\bibliographystyle{IEEEtran}
\bibliography{IEEEabrv,cmo}

\end{document}

%% file: introduction.tex

\section{Introduction}
\label{sec:introduction}

The constrained multi-objective optimization problem (CMOP) considered in this paper is defined as:
\begin{equation}
\begin{array}{l l}
\mathrm{minimize} \quad \mathbf{F}(\mathbf{x})=(f_{1}(\mathbf{x}),\cdots,f_{m}(\mathbf{x}))^{T}\\
\mathrm{subject\ to} \quad g_j(\mathbf{x})\geq a_j,\quad j=1,\cdots,q\\
\mathrm{\ } \quad\quad\quad\quad\quad h_j(\mathbf{x})=b_j,\quad j=q+1,\cdots,\ell\\
\mathrm{\ } \quad\quad\quad\quad\quad \mathbf{x} \in\Omega
\end{array}
\label{MOP}
\end{equation} 
where $\mathbf{x}=(x_1,\ldots,x_n)^T$ s a candidate solution, and $\Omega=[x_i^L,x_i^U]^n\subseteq\mathbb{R}^n$ defines the search (or decision variable) space. $\mathbf{F}:\Omega\rightarrow\mathbb{R}^m$ constitutes $m$ conflicting objective functions, and $\mathbb{R}^m$ is the objective space. $g_j(\mathbf{x})$ and $h_j(\mathbf{x})$ are the $j$-th inequality and equality constraints respectively. For a CMOP, the degree of constraint violation of $\mathbf{x}$ at the $j$-th constraint is calculated as~\cite{Deb01}:
\begin{equation}
    c_j(\mathbf{x})=
    \begin{cases}
        \langle g_j(\mathbf{x})/a_j-1\rangle, & j=1,\cdots,q \\
        \langle |h_j(\mathbf{x})/b_j-1|-\epsilon\rangle, & j=q+1,\cdots,\ell
    \end{cases}
\end{equation}
where $\epsilon$ is a relax term for the equality constraint, and $\langle\alpha\rangle$ returns 0 if $\alpha\geq 0$ otherwise it returns the negative of $\alpha$. The constraint violation value of $\mathbf{x}$ is calculated as:
\begin{equation}
	CV(\mathbf{x})=\sum_{j=1}^{\ell}c_j(\mathbf{x}),
    \label{eq:cv}
\end{equation}
$\mathbf{x}$ is feasible in case $CV(\mathbf{x})=0$; otherwise $\mathbf{x}$ is infeasible. Given two feasible solutions $\mathbf{x}^1$, $\mathbf{x}^2\in\Omega$, we said that $\mathbf{x}^1$ dominates $\mathbf{x}^2$ (denoted as $\mathbf{x}\preceq\mathbf{x}^2$) in case $\mathbf{F}(\mathbf{x}^1)$ is not worse than $\mathbf{F}(\mathbf{x}^2)$ in any individual objective and it at least has one better objective. A solution $\mathbf{x}^{\ast}$ is Pareto-optimal with respect to (\ref{MOP}) in case $\nexists\mathbf{x}\in\Omega$ such that $\mathbf{x}\preceq\mathbf{x}^{\ast}$. The set of all Pareto-optimal solutions is called the Pareto set (PS). Accordingly, $PF=\{\mathbf{F}(\mathbf{x})|\mathbf{x}\in PS\}$ is called the Pareto front (PF).

Since evolutionary algorithm (EA) is able to approximate a population of non-dominated solutions, which portray the trade-offs among conflicting objectives, in a single run, it has been recognized as a major approach for multi-objective optimization. Over the past two decades, much effort has been devoted to developing evolutionary multi-objective optimization (EMO) algorithms, e.g., elitist non-dominated sorting genetic algorithm (NSGA-II)~\cite{DebAPM02,LiKWCR12,LiKWTM13,LiDZZ17}, indicator-based EA~\cite{ZitzlerK04,BeumeNE07,LiKCLZS12} and multi-objective EA based on decomposition~\cite{ZhangL07,LiFKZ14,LiZKLW14,WuKZLWL15,WuLKZZ17}. Nevertheless, although most, if not all, real-life optimization scenarios have various constraints by nature, it is surprising that the research on constraint handling is lukewarm in the EMO community~\cite{WoldesenbetYT09}, comparing to algorithms designed for the unconstrained scenarios. 

Generally speaking, \textit{convergence}, \textit{diversity} and \textit{feasibility} are three basic issues for CMO. Most, if not all, currently prevalent constraint handling techniques at first tend to push a population toward the feasible region as much as possible, before considering the balance between convergence and diversity within the feasible region. This might lead to the population being stuck at some local optimal or local feasible regions, especially when the feasible regions are narrow and/or disparately distributed in the search space.

In this paper, we propose a two-archive EA, denoted as C-TAEA, for solving CMOPs. Specifically, we simultaneously maintain two co-evolving and complementary populations: one is denoted as convergence archive (CA); while the other is denoted as diversity archive (DA). The main characteristics of C-TAEA are delineated as follows:
\begin{itemize}
\item As the name suggests, the CA is the driving force to maintain the convergence and feasibility of the evolution process. It provides a consistent selection pressure toward the PF.
\item In contrast, without considering the feasibility, the DA mainly tends to maintain the convergence and diversity of the evolution process. In particular, the DA explores the areas that have not been exploited by the CA. This not only improves the population diversity of the CA within the currently investigating feasible region, but also helps jump over the local optima or local feasible regions.
\item To leverage the complementary effect and the elite information of these two co-evolving populations, we develop a restricted mating selection mechanism that selects the appropriate mating parents form the CA and DA separately according to their evolution status.
\end{itemize}

We admit that the two-archive or multi-population strategy is not a brand new technique in the EMO literature. For example, \cite{PraditwongY06,LiLTY14,WangJY15} developed several two-archive EMO algorithms that use two \lq\lq conceptually\rq\rq\ complementary populations to strike the balance between convergence and diversity of the evolutionary process. Li \textit{et al}.~\cite{LiKD15} developed a dual-population paradigm that combines the strengths of decomposition- and Pareto-based selection mechanisms. In this paper, we would like to, for the first time, explore the potential advantages of the two-archive strategy for CMOPs.

The rest of this paper is organized as follows. \pref{sec:preliminaries} briefly overviews the state-of-the-art evolutionary approaches developed for CMOPs and then elicits our motivations. \pref{sec:proposal} describes the technical details of the proposed algorithm step by step. Afterwards, in~\pref{sec:settings} and~\pref{sec:experiments}, the effectiveness and competitiveness of the proposed algorithm are empirically investigated and compared with five state-of-the-art constrained EMO algorithms on various benchmark problems. Finally, \pref{sec:conclusions} concludes with a summary and ideas for future directions.

%% file: related.tex

\section{Preliminaries}
\label{sec:preliminaries}

In this section, we first briefly review some recent developments of constraint handling techniques in the EMO community. Afterwards, we will give our motivations based on some examples.

\subsection{Literature Review}
\label{sec:related}

Generally speaking, the ideas of the existing constraint handling techniques in multi-objective optimization can be divided into the following three categories.

The first category is mainly driven by the feasibility information where feasible solutions are always granted a higher priority to survive to the next iteration. As early as the 90s, Fonseca and Flemming~\cite{FonsecaF98} developed a unified framework for solving MOPs with multiple constraints. In particular, they assign a higher priority to constraints than to objective functions. This results in a prioritization of the search for feasible solutions over optimal solutions. In~\cite{CoelloC99}, Coello Coello and Christiansen proposed a na\"ive constraint handling method that simply ignores the infeasible solutions. Although this method is easy to implement, it suffers the loss of selection pressure when tackling problems with a narrow feasible region. In particular, this algorithm will have no selection pressure when the population is filled with infeasible solutions. In~\cite{DebAPM02}, Deb \textit{et al}. developed a constrained dominance relation for CMO. Specifically, a solution $\mathbf{x}^1$ is said to constrained dominate another one $\mathbf{x}^2$ if: 1) $\mathbf{x}^1$ is feasible while $\mathbf{x}^2$ is not; 2) both of them are infeasible and $CV(\mathbf{x}^1)<CV(\mathbf{x}^2)$; 3) or both of them are feasible and $\mathbf{x}^1\prec\mathbf{x}^2$. By simply replacing the Pareto dominance relation with this constrained dominance relation, the state-of-the-art NSGA-II and NSGA-III~\cite{JainD14} can be readily used to tackle CMOPs. Borrowing the similar idea, several MOEA/D variants~\cite{JanZ10,JainD14,ChengJOS16} use the CV as an alternative criterion in the subproblem update procedure. Different from~\cite{DebAPM02}, Oyama \textit{et al}.~\cite{OyamaSF07} modified the Pareto dominance relation so that solutions who violate fewer number of constraints are preferred. To improve the interpretability of infeasible solutions, Takahama \textit{et al}.~\cite{TakahamaS12} and Mart\'inez \textit{et al}.~\cite{MartinezC14} proposed an $\epsilon$-constraint dominance relation where two solutions violate constraints equally in case the difference of their CVs is smaller than a threshold $\epsilon$. In particular, this threshold can be adaptively tuned according to the ratio of feasible solutions in the population. In~\cite{AsafuddoulaRS15}, Asafuddoula \textit{et al}. proposed an adaptive constraint handling method that treats infeasible solutions as feasible ones in case their CVs are less than a threshold. Analogously, Fan \textit{et al}.~\cite{FanLCHLL16} developed an angle-based constrained dominance principle by which two infeasible solutions are regarded as non-dominated from each other when their angle is larger than a threshold.

The second category aims at balancing the trade-off between convergence and feasibility during the search process. In~\cite{JimenezGSD02}, Jim\'enez \textit{et al}. proposed a min-max formulation that drives feasible solutions to evolve toward optimality and drives infeasible solutions to evolve toward feasibility. In~\cite{RayTS01}, Ray \textit{et al}. suggested a Ray-Tai-Seow algorithm that uses three different methods to compare and rank non-dominated solutions. Specifically, the first ranking procedure is conducted by sorting the objective values; the second one is performed according to different constraints; while the last one is based on a combination of objective values and constraints. Based on the same rigour, Young~\cite{Young05} proposed a constrained dominance relation that compares solutions according to the blended rank from both the objective space and the constraint space. A similar approach is developed by Angantyr \textit{et al}.~\cite{AngantyrAA03} that uses the weighted average rank of the ranks in both the objective space and the constraint space. By transforming each of the original objective functions of a CMOP into the sum of the distance measure and penalty function, \cite{WoldesenbetYT09} developed a new constraint handling technique for CMO. In particular, the modified objective functions are used in the non-dominated sorting procedure of NSGA-II to facilitate the search of optimal solutions in both feasible and infeasible regions. To improve the population diversity, Li \textit{et al}.~\cite{LiDZK15} developed a method that preserves infeasible solutions in case they are in the isolated regions. More recently, Ning \textit{et al}.~\cite{NingGYWWZ17} proposed a constrained non-dominated sorting method where each solution is assigned a constrained non-domination rank based on its Pareto rank and constraint rank.

The last category tries to repair the infeasible solutions and thus drives them toward the feasible region. For example, Harada \textit{et al}.~\cite{HaradaSOK06} proposed a so-called Pareto descent repair operator that explores possible feasible solutions around infeasible solutions in the constraint space. However, the gradient information is usually unavailable in practice. In~\cite{SinghRS10}, Singh \textit{et al}. suggested to use simulated annealing to accelerate the progress of movements from infeasible solutions toward feasible ones. Jiao \textit{et al}.~\cite{JiaoLSL14} developed a feasible-guiding strategy in which the feasible direction is defined as a vector starting from an infeasible solution and ending up with its nearest feasible solution. Afterwards, infeasible solutions are guided toward the feasible region by leveraging the information provided by the feasible direction.

\subsection{Challenges to Existing Constraint Handling Techniques}
\label{sec:motivations}

From the above literature review, we find that most, if not all, constraint handling techniques in multi-objective optimization overly emphasize the importance of feasibility, whereas they rarely consider the balance among convergence, diversity and feasibility simultaneously. This can lead to an ineffective search when encountering complex constraints.



\begin{figure}[htbp]
\centering
\includegraphics[width=.8\linewidth]{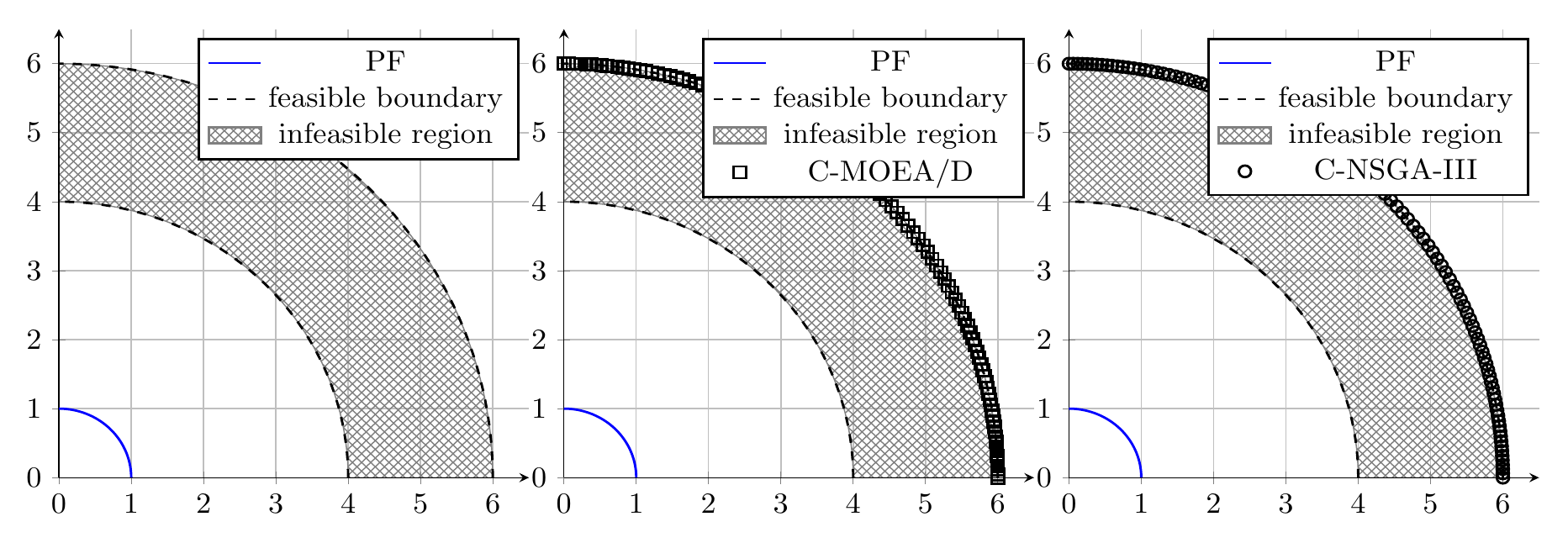}
\caption{Comparative results on the two-objective C1-DTLZ3.}
\label{fig:C1}
\end{figure}

Let us first consider a test problem C1-DTLZ3 defined in~\cite{JainD14}, where the objective functions are the same as the classic DTLZ3 problem~\cite{DebTLZ05} while the constraint is defined as:
\begin{equation}
c(\mathbf{x})=(\sum_{i=1}^mf_i(\mathbf{x})^2-16)(\sum_{i=1}^mf_i(\mathbf{x})^2-r^2)\geq 0,
\end{equation}
\pref{fig:C1} shows a two-objective example where $r$ is set to 6. From this figure, we can see that the feasible region of this test problem is intersected by an infeasible ribbon. In addition, within this infeasible region, the CV of a solution increases when it moves away from the feasible boundary, and decreases otherwise. Therefore, it is not difficult to infer that a feasibility-driven strategy will be easily trapped in the outermost feasible boundary. To validate this assertion, we employ the state-of-the-art C-MOEA/D and C-NSGA-III~\cite{JainD14} as the benchmark algorithms where the corresponding parameters are set the same as~\cite{JainD14}. As shown in~\pref{fig:C1}, solutions found by both algorithms are stuck in the outermost feasible boundary after 1,000 generations.

\begin{figure}[htbp]
\centering
\includegraphics[width=.8\linewidth]{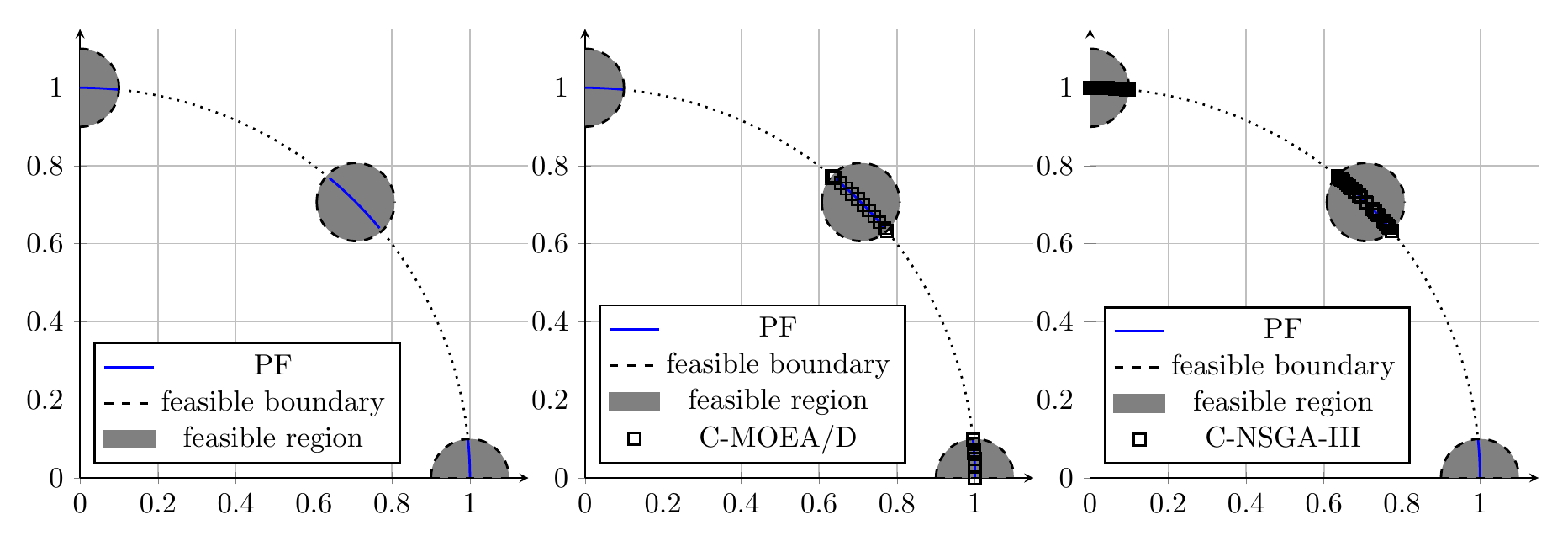}
\caption{Comparative results on the two-objective C2-DTLZ2.}
\label{fig:C2}
\end{figure}

Let us consider another test problem C2-DTLZ2 defined in~\cite{JainD14}, where the objective functions are the same as the classic DTLZ2 problem~\cite{DebTLZ05} while the constraint is defined as:
\begin{align}
c(\mathbf{x})&=\max\bigg\{\max_{i=1}^m\Big[(f_i(\mathbf{x})-1)^2+\sum_{j=1,j\neq i}^mf_j^2-r^2\Big],\nonumber\\
&\Big[\sum_{i=1}^m(f_i(\mathbf{x})-\frac{1}{\sqrt{m}})^2-r^2\Big]\bigg\},
\end{align}
\pref{fig:C2} gives an example in the two-objective scenario, where three feasible regions are sparsely located on the PF. If the size of each feasible region is small, a feasibility-driven strategy will be easily trapped in some, not all, of the feasible regions. Furthermore, it is highly likely that none of the weight vectors used in the state-of-the-art decomposition-based EMO algorithms, e.g., C-MOEA/D and C-NSGA-III, cross these feasible regions if their sizes are sufficiently small. In this case, the decomposition-based EMO algorithms will be struggled to find feasible solutions. The results shown in~\pref{fig:C2} fully validate our assertions, where neither C-MOEA/D nor C-NSGA-III can find Pareto-optimal solutions on all three feasible regions when we set $r$ to be a relatively small value, say 0.1.

Based on these discussions, we find that an excessive use of the feasibility information can restrict the search ability of a constrained EMO algorithm. In~\pref{sec:proposal}, we will demonstrate how to use a two-archive strategy to balance the convergence, diversity and feasibility simultaneously in the entire search space. In particular, we find that an appropriate use of the infeasibility information can help to resolve the dilemma between exploration versus exploitation.

%% file: proposal.tex

\section{Proposed Algorithm}
\label{sec:proposal}

\begin{figure*}[htbp]
\centering
\includegraphics[width=.9\linewidth]{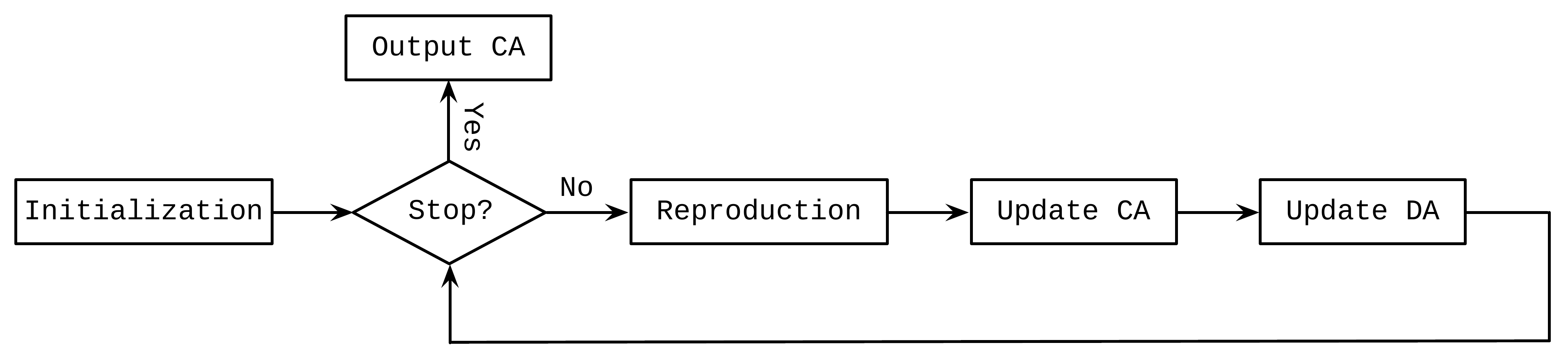}
\caption{Flow chart of C-TAEA.}
\label{fig:flowchart}
\end{figure*}

The general flow chart of our proposed C-TAEA is given in~\pref{fig:flowchart}. As its name suggests, C-TAEA maintains two co-evolving archives, named CA and DA, each of which has the same and fixed size $N$. Specifically, CA, as the main force, is mainly responsible for driving the population toward the feasible region and approximating the PF; DA, as a complement, is mainly used to explore the areas under-exploited by the CA. It is worth noting that, to provide as much diversified information as possible, the update of the DA does not take the feasibility information into account. During the reproduction process, mating parents are separately selected from the CA and DA according to their evolution status, as described in~\pref{sec:reproduction}. Afterwards, the offspring are used to update the CA and DA according to the mechanisms described in~\pref{sec:updateCA} and~\pref{sec:updateDA} separately.

\subsection{Density Estimation Method}
\label{sec:density}

\begin{algorithm}[htbp]
\KwIn{Solution set $\mathrm{S}$, weight vector set $\mathrm{W}$}
\KwOut{Subregions $\Delta^1,\cdots,\Delta^N$}
$\Delta^1\leftarrow\emptyset,\cdots,\Delta^N\leftarrow\emptyset$;\\
\ForEach{$\mathbf{x}\in\mathrm{S}$}{
    \ForEach{$\mathbf{w}\in\mathrm{W}$}{
        Compute $d^{\perp}(\mathbf{x},\mathbf{w})=\mathbf{x}-\mathbf{w}^T\mathbf{x}/\|\mathbf{w}\|$;
    }
    $k\leftarrow\argmin\limits_{\mathbf{w}\in W}d^{\perp}(\mathbf{x},\mathbf{w})$;\\
    $\Delta^k\leftarrow\Delta^k\bigcup\{\mathbf{x}\}$;
}

\Return $\Delta^1,\cdots,\Delta^N$

\caption{Association Procedure}
\label{alg:association}
\end{algorithm}

Before explaning the update mechanisms of the CA and DA in C-TAEA, we first introduce the density estimation method that is useful for both cases. To facilitate the density estimation, we borrow the idea from~\cite{LiuGZ14} to divide the objective space into $N$ subregions, each of which is represented by a unique weight vector on the canonical simplex. In particular, we employ our previously developed weight vector generation method~\cite{LiDZK15}, which is scalable to the many-objective scenarios, to sample a set of uniformly distributed weight vectors, i.e., $\mathrm{W}=\{\mathbf{w}^1,\cdots,\mathbf{w}^N\}$. Specifically, a subregion $\Delta^i$, where $i\in\{1,\cdots,N\}$, is defined as:
\begin{equation}
\Delta^i=\{\mathbf{F}(\mathbf{x})\in\mathbb{R}^m|\langle\mathbf{F}(\mathbf{x}),\mathbf{w}^i\rangle\leq\langle\mathbf{F}(\mathbf{x}),\mathbf{w}^j\rangle\},
\end{equation} 
where $j\in\{1,\cdots,N\}$ and $\langle\mathbf{F}(\mathbf{x}),\mathbf{w}\rangle$ is the acute angle between $\mathbf{F}(\mathbf{x})$ and the reference line formed by the origin and $\mathbf{F}(\mathbf{x})$. After the setup of subregions, each solution $\mathbf{x}$ of a population is associated with a unique subregion whose index is determined as:
\begin{equation}
k=\argmin\limits_{i\in\{1,\cdots,N\}}\langle\mathbf{\overline{F}}(\mathbf{x}),\mathbf{w}^i\rangle,
\end{equation}
where $\mathbf{\overline{F}}(\mathbf{x},t)$ is the normalized objective vector of $\mathbf{x}$, and its $i$-th objective function is calculated as:
\begin{equation}
\overline{f}_i(\mathbf{x})=\frac{f_i(\mathbf{x})-z^{\ast}_i}{z^{nad}_i-z^{\ast}_i},
\end{equation}
where $i\in\{1,\cdots,m\}$, $\mathbf{z}^{\ast}$ and $\mathbf{z}^{nad}$ are respectively the estimated ideal and nadir points, where $z^{\ast}_i=\min\limits_{\mathbf{x}\in S}f_i(\mathbf{x})$ and $z^{nad}_i=\max\limits_{\mathbf{x}\in S}f_i(\mathbf{x})$. The pseudo-code of this association procedure is given in~\pref{alg:association}. After associating solutions with subregions, the density of a subregion is counted as the number of its associated solutions.

\begin{algorithm}[htbp]
\KwIn{$\mathrm{CA}$, offspring population $\mathrm{Q}$, weight vector set $\mathrm{W}$}
\KwOut{Updated $\mathrm{CA}$}

$\mathrm{S}\leftarrow\emptyset$, $\mathrm{S}_c\leftarrow\emptyset$, $i\leftarrow 1$, $\mathrm{H}_c\leftarrow\mathrm{CA}\bigcup\mathrm{Q}$;\\
    \ForEach{$\mathbf{x}\in\mathrm{H}_c$}{
        \If{$CV(\mathbf{x})=0$}{
            $\mathrm{S}_c\leftarrow\mathrm{S}_c\bigcup\{\mathbf{x}\}$;
        }
    }

\uIf{$|\mathrm{S}_c|=N$}{
    $\mathrm{CA}\leftarrow\mathrm{S}_c$;
}\uElseIf{$|\mathrm{S}_c|>N$}{
    Use non-dominated sorting to divide $\mathrm{S}_c$ into $\{F_1,F_2,\cdots\}$ based on the MOP defined in~(\ref{MOP});\\
    \While{$|\mathrm{S}|<N$}{
        $\mathrm{S}\leftarrow\mathrm{S}\bigcup F_i$, $i\leftarrow i+1$;\\
    }
    \If{$|\mathrm{S}|>N$}{
        \ForEach{$\mathbf{x}\in\mathrm{S}$}{
            $\overline{\mathbf{F}}_k(\mathbf{x})=\frac{\mathbf{F}(\mathbf{x})-\mathbf{z}^{\ast}}{\mathbf{z}^{nad}-\mathbf{z}^{\ast}}$;
        }
        $\{\Delta^1,\cdots,\Delta^N\}\leftarrow$ \texttt{Association}$(\mathrm{S},\mathrm{W})$;\\
        \While{$|\mathrm{S}|>N$}{
            Find the most crowded subregion $\Delta^i$;\\
            \ForEach{$\mathbf{x}\in\Delta^i$}{
            	$dist(\mathbf{x})\leftarrow\min\limits_{\mathbf{x}'\in\Delta^i,\mathbf{x}\neq\mathbf{x}'}\|\mathbf{x}-\mathbf{x}'\|$;
            }
            $\mathrm{S}_t\leftarrow\argmin\limits_{\mathbf{x}\in\Delta^i}\{dist(\mathbf{x})\}$;\\
            $\mathbf{x}^w\leftarrow\argmax\limits_{\mathbf{x}\in\mathrm{S}_t}\{g^{tch}(\mathbf{x}|\mathbf{w}^i,\mathbf{z}^{\ast})\}$;\\
            $\mathrm{S}\leftarrow\mathrm{S}\setminus\{\mathbf{x}^w\}$;
        }        
    }
    $\mathrm{CA}\leftarrow\mathrm{S}$;
}\Else{
    $\mathrm{S}_I\leftarrow\mathrm{H}_c\setminus\mathrm{S}_c$;\\
    Use non-dominated sorting to divide $\mathrm{S}_I$ into $\{F_1,F_2,\cdots\}$ based on the MOP defined in~(\ref{newMOP});\\
    \While{$|\mathrm{S}_c|<N$}{
    $\mathrm{S}\leftarrow\mathrm{S}\bigcup F_i$, $i\leftarrow i+1$;\\
    }
    \While{$|\mathrm{S}|>N$}{
		$\mathbf{x}^w\leftarrow\argmax\limits_{\mathbf{x}\in F_{i-1}}\{CV(\mathbf{x})\}$, $\mathrm{S}\leftarrow\mathrm{S}\setminus\{\mathbf{x}^w\}$;
    }
    $\mathrm{CA}\leftarrow\mathrm{S}$;
} 
\Return $\mathrm{CA}$

\caption{Update Mechanism of CA}
\label{alg:updateCA}
\end{algorithm}

\subsection{Update Mechanism of the CA}
\label{sec:updateCA}

The effect of the CA is similar to the other constrained EMO algorithms in the literature. It first pushes the population toward the feasible region as much as possible, then it tries to balance the convergence and diversity within the feasible region. The pseudo-code of the update mechanism of the CA is given in~\pref{alg:updateCA}. Specifically, we first form a hybrid population $\mathrm{H}_c$, a combination of the CA and the offspring population $\mathrm{Q}$. Feasible solutions in $\mathrm{H}_c$ are chosen into a temporary archive $\mathrm{S}_c$ (lines 3 to 5 of~\pref{alg:updateCA}). Afterwards, the follow-up procedure depends on the size of $\mathrm{S}_c$:
\begin{itemize}
\item If the size of $\mathrm{S}_c$ equals $N$ (i.e., the predefined size of the CA), it is directly used as the new CA and this update procedure terminates (lines 6 and 7 of~\pref{alg:updateCA}).
\item If $|\mathrm{S}_c|>N$, we use the fast non-dominated sorting method~\cite{DebAPM02} to divide $\mathrm{S}_c$ into several non-domination levels, i.e., $F_1$, $F_2$, and so on. Starting from $F_1$, each non-domination level is sequentially chosen to construct a temporary archive $\mathrm{S}$ until its size equals or for the first time exceeds $N$ (lines 9 to 11 of~\pref{alg:updateCA}). If we denote the last acceptable non-domination level as $F_l$, solutions belonging to $F_{l+1}$ onwards are exempt from further consideration. Note that $\mathrm{S}$ can be used as the new CA if its size equals $N$; otherwise we associate each solution in $\mathrm{S}$ with its corresponding subregion and calculate $\mathrm{S}$'s density information afterwards. Iteratively, a worst solution from the most crowded subregion (tie is broken randomly) is trimmed one at a time until $\mathrm{S}$'s size equals $N$ (line 11 to 21 of~\pref{alg:updateCA}). Note that, to improve the population diversity within a subregion, we propose the following process to identify the worst solution $\mathbf{x}^w$. First, we calculate the distance between each solution $\mathbf{x}$ in $\Delta^i$ and its nearest neighbor:
\begin{equation}
dist(\mathbf{x})=\min_{\mathbf{x}'\in\Delta^i,\mathbf{x}\neq\mathbf{x}'}\|\mathbf{x}-\mathbf{x}'\|,
\end{equation} 
where $\|\cdot\|$ indicates the $\ell_2$-norm. Afterwards, the solutions having the smallest distance are stored in a temporary archive $\mathrm{S}_t$, while $\mathbf{x}^w$ is defined as
\begin{equation}
\mathbf{x}^w=\argmax\limits_{\mathbf{x}\in\mathrm{S}_t}\{g^{tch}(\mathbf{x}|\mathbf{w}^i,\mathbf{z}^{\ast})\},
\label{eq:fitness}
\end{equation}
where
\begin{equation}
g^{tch}(\mathbf{x}|\mathbf{w}^i,\mathbf{z}^{\ast})=\max\limits_{1\leq j\leq m}\{|f_j(\mathbf{x})-z^{\ast}_j|/w_j^i\}.
\label{eq:fitness}
\end{equation}
\item Otherwise, if the feasible solutions in $\mathrm{H}_c$ do not fill the new CA ($|\mathrm{S}_c|<N$), we formulate a new bi-objective optimization problem as follows:
\begin{equation}
\begin{array}{l l}
\mathrm{minimize} \quad \mathbf{F}(\mathbf{x})=(f_1(\mathbf{x}),f_2(\mathbf{x}))^{T}\\
\mathrm{where } \quad\quad
\begin{cases} 
f_1(\mathbf{x})=CV(\mathbf{x})\\
f_2(\mathbf{x})=g^{tch}(\mathbf{x}|\mathbf{w}^i,\mathbf{z}^{\ast})
\end{cases} 
\end{array}
\label{newMOP}
\end{equation}
Based on~(\ref{newMOP}), we use the fast non-dominated sorting method to divide the infeasible solutions in $\mathrm{H}_c$ into several non-domination levels (lines 24 and 25 of~\pref{alg:updateCA}). Solutions in the first several levels have a higher priority to survive into the new CA. Exceeded solutions are trimmed according to their CVs, i.e., the solution having a larger CV is trimmed at first (lines 28 to 29 of~\pref{alg:updateCA}). These operations tend to further balance the convergence, diversity and feasibility.
\end{itemize}

\subsection{Update Mechanism of the DA}
\label{sec:updateDA}

Different from the CA, the DA aims at providing as much diversified solutions as possible. In particular, its update mechanism has two characteristics: 1) it does not take the constraint violation into consideration; 2) it takes the up to date CA as a reference set so that it complements the behavior of the CA by exploring its under-exploited areas. The pseudo-code of this update procedure is presented in~\pref{alg:updateDA}. Specifically, similar to~\pref{sec:updateCA}, we at first combine the DA with the offspring population $\mathrm{Q}$ to form a hybrid population $\mathrm{H}_d$. Then, we separately associate each solution in $\mathrm{H}_d$ and the up to date CA with its corresponding subregion according to the method introduced in~\pref{sec:density} (lines 1 to 3 of~\pref{alg:updateDA}). Afterwards, we iteratively investigate each subregion and decide the survival of solutions in $\mathrm{H}_d$ to the new DA. In particular, at the $\mathrm{itr}$-th iteration, at most $\mathrm{itr}$ solutions, including those in the CA and $\mathrm{H}_d$, can survive in each subregion. In other words, for the currently investigating subregion, say $\Delta^i,i\in\{1,\cdots,N\}$, if there already exists $\mathrm{itr}$ solutions in CA at $\Delta^i$, no solution in $\mathrm{H}_d$ will be considered to survive at $\Delta^i$ during this iteration. Otherwise, the best non-dominated solutions in $\mathrm{H}_d$ associated with $\Delta^i$, denoted as $\mathbf{O}^i$, will be chosen to survive to the new DA (lines 10 to 12 of~\pref{alg:updateDA}). Here the best solution $\mathbf{x}^b$ is identified as:
\begin{equation}
    \mathbf{x}^b=\argmin\limits_{\mathbf{x}\in\mathrm{O}^i}\{g^{tch}(\mathbf{x}|\mathbf{w}^i,\mathbf{z}^{\ast})\}.
\end{equation}
This iterative investigation continues till the DA is filled.

\begin{algorithm}[t]
\KwIn{$\mathrm{CA}$, $\mathrm{DA}$, offspring population $\mathrm{Q}$, weight vector set $\mathrm{W}$}
\KwOut{Updated $\mathrm{DA}$}

$\mathrm{S}\leftarrow\emptyset$, $i\leftarrow 1$, $\mathrm{H}_d\leftarrow\mathrm{DA}\bigcup\mathrm{Q}$;\\

$\{\Delta^1_d,\cdots,\Delta^N_d\}\leftarrow$ \texttt{Association}$(\mathrm{H}_d,\mathrm{W})$;\\
$\{\Delta^1_c,\cdots,\Delta^N_c\}\leftarrow$ \texttt{Association}$(\mathrm{CA},\mathrm{W})$;\\

$\mathsf{itr}\leftarrow 1$;\\
\While{$|\mathrm{S}|\leq N$}{
    \For{$i\leftarrow 1$ \KwTo $N$}{
		\If{$|\Delta^i_c|<\mathrm{itr}$}{
        	\For{$i\leftarrow 1$ \KwTo $\mathrm{itr}-|\Delta^i_c|$}{
				\uIf{$\Delta^i_d\neq\emptyset$}{
					$\mathrm{O}^i\leftarrow$ non-dominated solutions in $\Delta^i_d$;\\
					$\mathbf{x}^b\leftarrow\argmin\limits_{\mathbf{x}\in\mathrm{O}^i}\{g^{tch}(\mathbf{x}|\mathbf{w}^c,\mathbf{z}^{\ast})\}$;\\
					$\Delta^i_d\leftarrow\Delta^i_d\setminus\{\mathbf{x}^b\}$, $\mathrm{S}\leftarrow\mathrm{S}\bigcup\{\mathbf{x}^b\}$;
				}\Else{
				\textbf{break};
				}	
			}
        }
    }
    $\mathsf{itr}\leftarrow \mathsf{itr}+1$;
}
$\mathrm{DA}\leftarrow\mathrm{S}$;

\Return $\mathrm{DA}$

\caption{Update Mechanism of the DA}
\label{alg:updateDA}
\end{algorithm}

\subsection{Offspring Reproduction}
\label{sec:reproduction}

The interaction and collaboration between two co-evolving archives is a vital step in C-TAEA. Apart from the complementary behavior of the update mechanisms of the CA and DA, the other contributing factor for this collaboration is the restricted mating selection. Generally speaking, its major purpose is to leverage the elite information from both archives for offspring reproduction. \pref{alg:mating} provides the pseudo code of this restricted mating selection procedure. Specifically, we first combine the CA and the DA into a composite set $\mathrm{H}_m$. Afterwards, we separately evaluate the proportion of non-dominated solutions of the CA and the DA in $\mathrm{H}_m$ (lines 2 and 3 of~\pref{alg:mating}). If $\rho_c>\rho_d$, it means that the convergence status of the CA is better than that of the DA. Accordingly, the first mating parent is chosen from the CA; otherwise, it comes from the DA (lines 4 to 7 of~\pref{alg:mating}). As for the other mating parent, whether it is chosen from the CA or the DA depends on the proportion of non-dominated solutions (lines 8 to 11 of~\pref{alg:mating}). In other words, the higher proportion of non-dominated solutions, the larger chance to be chosen as the mating pool. As shown in lines 5 to 11 of~\pref{alg:mating}, we use a binary tournament selection to choose a mating parent. As shown in~\pref{alg:tournament}, this tournament selection procedure is feasibility-driven. Specifically, if the randomly selected candidates are all feasible, they are chosen based on the Pareto dominance; if only one of them is feasible, the feasible one will be chosen; otherwise, the mating parent is chosen in a random manner. Once the mating parents are chosen, we use the popular simulated binary crossover~\cite{DebA94} and the polynomial mutation~\cite{DebG96} for offspring reproduction. In principle, any other reproduction operator can be readily applied with a minor modification.

\begin{algorithm}[t]
\KwIn{$\mathrm{CA}$, $\mathrm{DA}$}
\KwOut{Mating parents $\mathbf{p}_1$, $\mathbf{p}_2$}

$\mathrm{H}_m\leftarrow\mathrm{CA}\bigcup\mathrm{DA}$;\\
$\rho_c\leftarrow$proportion of non-dominated solution of $\mathrm{CA}$ in $\mathrm{H}_m$;\\
$\rho_d\leftarrow$proportion of non-dominated solution of $\mathrm{DA}$ in $\mathrm{H}_m$;\\ 
\uIf{$\rho_c>\rho_d$}{
    $\mathbf{p}_1\leftarrow$\texttt{TournamentSelection}$(\mathrm{CA})$;
}\Else{
    $\mathbf{p}_1\leftarrow$\texttt{TournamentSelection}$(\mathrm{DA})$;
}
\uIf{$\texttt{rand}<\rho_c$}{
    $\mathbf{p}_2\leftarrow$\texttt{TournamentSelection}$(\mathrm{CA})$;
}\Else{
    $\mathbf{p}_2\leftarrow$\texttt{TournamentSelection}$(\mathrm{DA})$; 
}
\textbf{return} $\mathbf{p}_1$, $\mathbf{p}_2$

\caption{Restricted Mating Selection}
\label{alg:mating}
\end{algorithm}

%
%

\begin{algorithm}[t]
\KwIn{Solution set $S$}
\KwOut{Mating parent $\overline{\mathbf{x}}$}

Randomly pick two solutions $\mathbf{x}^1$ and $\mathbf{x}^2$ from $S$;\\
\uIf{$\mathbf{x}^1$ and $\mathbf{x}^2$ are feasible}{
	\uIf{$\mathbf{x}^1\preceq\mathbf{x}^2$}{
		$\overline{\mathbf{x}}\leftarrow\mathbf{x}^1$;
	}\uElseIf{$\mathbf{x}^2\preceq\mathbf{x}^1$}{
		$\overline{\mathbf{x}}\leftarrow\mathbf{x}^2$;
	}\Else{
		$\overline{\mathbf{x}}\leftarrow$Randomly pick one from $\mathbf{x}^1$ and $\mathbf{x}^2$;
	}
}\uElseIf{Only one solution is feasible}{
    $\overline{\mathbf{x}}\leftarrow$feasible one from $\mathbf{x}^1$ and $\mathbf{x}^2$;
}\Else{
		$\overline{\mathbf{x}}\leftarrow$Randomly pick one from $\mathbf{x}^1$ and $\mathbf{x}^2$;
}
\textbf{return} $\overline{\mathbf{x}}$
\caption{Tournament Selection}
\label{alg:tournament}
\end{algorithm}

%% file: settings.tex
\section{Experimental Setup}
\label{sec:settings}

Before discussing the empirical results, this section briefly introduces the benchmark problems, performance metrics and the state-of-the-art constrained EMO algorithms used for peer comparisons in our empirical studies.

\subsection{Benchmark Suite}
\label{sec:problems}

Five constrained test problems (i.e., C1-DTLZ1/DTLZ3, C2-DTLZ2 and C3-DTLZ1/DTLZ4) from~\cite{JainD14} and six newly proposed test problems (DC1-DTLZ1/DTLZ3, DC2-DTLZ2/DTLZ4 and DC3-DTLZ1/DTLZ4) are chosen to form the benchmark suite. All these test problems are scalable to any number of objectives, where we set $m\in\{3,5,8,10,15\}$ here. Detailed descriptions, including the mathematical definitions and properties, of these test problems are given in Section I of the supplementary document.

\subsection{Performance Metrics}
\label{sec:metrics}

Two widely used metrics are chosen to assess the performance of different algorithms. 
\begin{enumerate}
	\item\textit{Inverted Generational Distance} (IGD)~\cite{BosmanT03}: Given $P^{\ast}$ as a set of points uniformly sampled along the PF and $P$ as the set of solutions obtained from an EMO algorithm. The IGD value of $P$ is calculated as:
	\begin{equation}
	IGD(P,P^*)=\frac{\sum_{\mathbf{z}\in P^*}dist(\mathbf{z},P)}{|P^*|},
	\end{equation}
	where $dist(\mathbf{z},P)$ is the Euclidean distance between $\mathbf{z}$ and its nearest neighbor in $P$.
	\item\textit{Hypervolume} (HV)~\cite{ZitzlerT99}: Let $\mathbf{z}^r=(z^r_1,\cdots,z^r_m)^T$ be a worst point dominated by all the Pareto optimal objective vectors. The HV of $P$ is defined as the volume of the objective space dominated by solutions in $P$ and bounded by $\mathbf{z}^r$:
	\begin{equation}
	HV(P)=\textsf{VOL}(\bigcup \limits_{\mathbf{z}\in P} [z_1,z^r_1]\times\cdots\times [z_m,z^r_m]),
	\end{equation}
	where \textsf{VOL} indicates the Lebesgue measure.
\end{enumerate}
To calculate the IGD, we need to sample a sufficient number of points from the PF to form $P^{\ast}$. For C-DTLZ problem instances, we use the method developed in~\cite{LiDZK15} to fulfill this purpose. Before calculating the HV, we remove the solutions dominated by the $\mathbf{z}^r$, which is set as $(\underbrace{1.1,\cdots,1.1}_\text{$m$})^T$ in our empirical studies, except for C3-DTLZ4 where $\mathbf{z}^r=(\underbrace{2.1,\cdots,2.1}_\text{$m$})^T$. Note that both IGD and HV can evaluate the convergence and diversity simultaneously. A smaller IGD or a larger HV value indicates a better approximation to the PF. Each algorithm is independently run 51 times. The median and the interquartile range (IQR) of the IGD and HV values are presented in the corresponding tables. In particular, the best results are highlighted in boldface with a gray background. To have a statistically sound conclusion, we use the Wilcoxon's rank sum test at a significant level of 5\% to validate the significance of the better performance achieved by the proposed C-TAEA with respect to the other peer algorithms.

\subsection{EMO Algorithms Used for Comparisons}
\label{sec:peers}

Five state-of-the-art constrained EMO algorithms, i.e., C-MOEA/D, C-NSGA-III, C-MOEA/DD~\cite{LiDZK15}, I-DBEA~\cite{AsafuddoulaRS15} and CMOEA~\cite{WoldesenbetYT09}, are chosen for peer comparisons. All algorithms use the simulated binary crossover and the polynomial mutation for offspring generation. The termination criteria is a predefined number of function evaluations. Section II of the supplementary document briefly describes these peer algorithms and lists their corresponding parameter settings.

%% file: experiments.tex

\section{Empirical Studies}
\label{sec:experiments}

In this section, we discuss the empirical results on different benchmark problems separately.

\subsection{C-DTLZ Benchmark Suite}
\label{sec:CDTLZ}

\begin{table*}[htbp]
\centering 
\caption{Comparison results on IGD metric (median and IQR) for C-TAEA and the other peer algorithms on C-DTLZ Benchmark Suite}
\label{tab:CDTLZ-IGD}
\resizebox{\textwidth}{!}{ 
\begin{tabular}{c|c|c|c|c|c|c|c}
\hline
                         & $m$ & C-TAEA            & C-NSGA-III        & C-MOEA/D          & C-MOEA/DD         & I-DBEA           & CMOEA             \\ \hline
\multirow{5}{*}{C1-DTLZ1} & 3   & 2.069E-2(1.33E-5) & \bb{2.037E-2(7.06E-5)$^{\ddag}$} & 2.110E-2(3.17E-4)$^{\dag}$ & 2.116E-2(4.75E-4)$^{\dag}$ & 2.180E-2(6.03E-6)$^{\dag}$ & 2.140E-2(5.45E-4)$^{\dag}$ \\ \cline{2-8}
                         & 5   & \bb{5.278E-2(1.16E-3)} & 5.427E-2(1.62E-3)$^{\dag}$ & 5.294E-2(7.79E-5)$^{\dag}$ & 5.287E-2(1.81E-5) & 5.285E-2(6.62E-5) & 5.284E-2(1.97E-5)  \\ \cline{2-8}
                         & 8   & \bb{9.912E-2(1.60E-3)} & 1.009E-1(1.59E-3) & 1.006E-1(6.93E-4) & 1.024E-1(1.86E-3)$^{\dag}$ & 1.009E-1(5.30E-4) & 1.008E-1(5.76E-4)  \\ \cline{2-8}
                         & 10  & 1.061E-1(3.82E-3) & \bb{1.038E-1(8.86E-3)$^{\ddag}$} & 1.074E-1(7.81E-2) & 1.065e-1(9.08E-2) & 1.072E-1(7.87E-3) & 1.072E-1(3.39E-3) \\ \cline{2-8}
                         & 15  & \bb{2.233E-1(8.02E-4)} & 2.351E-1(3.40E-3)$^{\dag}$ & 2.608E-1(7.62E-3)$^{\dag}$ & 2.490E-1(6.53E-3)$^{\dag}$ & 2.506E-1(4.47E-3)$^{\dag}$ & 2.611E-1(7.25E-3)$^{\dag}$  \\ \hline\hline
\multirow{5}{*}{C1-DTLZ3} & 3  & \bb{5.661E-2(8.49E-3)} & 8.020E+0(4.22E-3)$^{\dag}$ & 8.007E+0(1.72E-3)$^{\dag}$ & 8.012E+0(1.08E-3)$^{\dag}$ & 8.013E+0(7.59E-3)$^{\dag}$ & 8.007E+0(2.07E-3)$^{\dag}$ \\ \cline{2-8}
                         & 5   & \bb{5.364E-1(9.03E-1)} & 1.162E+1(3.96E-2)$^{\dag}$ & 1.154E+1(4.41E-3)$^{\dag}$ & 1.155E+1(1.12E+1)$^{\dag}$ & 1.153E+1(4.79E-3)$^{\dag}$ & 1.154E+1(9.23E-3)$^{\dag}$ \\ \cline{2-8}
                         & 8   & \bb{4.115E-1(1.31E-2)} & 1.180E+1(8.59E-2)$^{\dag}$ & 1.160E+1(2.64E-3)$^{\dag}$ & 1.161E+1(4.47E-4)$^{\dag}$ & 1.160E+1(6.98E-3)$^{\dag}$ & 1.159E+1(1.84E-2)$^{\dag}$ \\ \cline{2-8}
                         & 10  & \bb{3.896E-1(8.75E-2)} & 1.430E+1(3.30E-2)$^{\dag}$ & 1.414E+1(1.93E-2)$^{\dag}$ & 1.414E+1(7.36E-3)$^{\dag}$ & 1.416E+1(6.11E-3)$^{\dag}$ & 1.412E+1(2.90E-2)$^{\dag}$ \\ \cline{2-8}
                         & 15  & \bb{8.749E-1(3.16E-2)} & 1.470E+1(5.33E-3)$^{\dag}$ & 1.466E+1(8.22E-2)$^{\dag}$ & 1.461E+1(4.30E-2)$^{\dag}$ & 1.463E+1(1.26E-2)$^{\dag}$ & 1.461E+1(6.31E-2)$^{\dag}$ \\ \hline\hline
\multirow{5}{*}{C2-DTLZ2} & 3  & \bb{1.594E-2(2.95E-3)} & 9.043E-1(1.25E-4)$^{\dag}$ & 9.069E-1(3.74E-1)$^{\dag}$ & 5.648E-1(3.67E-1)$^{\dag}$ & 9.069E-1(1.76E-3)$^{\dag}$ & 9.069E-1(1.05E-2)$^{\dag}$ \\ \cline{2-8}
                         & 5   & \bb{3.386E-1(1.46E-1)} & 1.068E+0(2.59E-5)$^{\dag}$ & 4.863E-1(5.93E-1)$^{\dag}$ & 1.069E+0(3.97E-2)$^{\dag}$ & 1.070E+0(1.54E-3)$^{\dag}$ & 1.074E+0(4.35E-3)$^{\dag}$ \\ \cline{2-8}
                         & 8   & \bb{1.310E-4(8.22E-4)} & 1.206E+0(1.25E-5)$^{\dag}$ & 1.220E+0(7.64E-3)$^{\dag}$ & 1.237E+0(2.27E-6)$^{\dag}$ & 1.051E+0(1.84E-1)$^{\dag}$ & 1.223E+0(6.64E-4)$^{\dag}$ \\ \cline{2-8}
                         & 10  & \bb{2.600E-5(1.03E-6)} & 1.241E+0(7.00E-6)$^{\dag}$ & 1.254E+0(5.57E-3)$^{\dag}$ & 1.273E+0(1.28E-5)$^{\dag}$ & 1.263E+0(1.46E-1)$^{\dag}$ & 1.257E+0(4.48E-3)$^{\dag}$ \\ \cline{2-8}
                         & 15  & \bb{5.658E-1(2.38E-3)} & 1.287E+0(3.34E-4)$^{\dag}$ & 1.317E+0(6.43E-2)$^{\dag}$ & 1.320E+0(7.21E-1)$^{\dag}$ & 1.315E+0(3.64E-2)$^{\dag}$ & 1.316E+0(3.79E-2)$^{\dag}$ \\ \hline\hline
\multirow{5}{*}{C3-DTLZ1} & 3  & \bb{4.311E-2(1.22E-4)} & 7.653E-2(1.40E-3)$^{\dag}$ & 4.344E-2(2.86E-2)$^{\dag}$ & 9.344E-2(1.98E-4)$^{\dag}$ & 4.435E-2(4.79E-3)$^{\dag}$ & 4.435E-2(1.22E-3)$^{\dag}$ \\ \cline{2-8}
                         & 5   & 1.073E-1(3.06E-5) & 1.124E-1(2.76E-3)$^{\dag}$ & \bb{1.073E-1(5.84E-5)} & 1.438E-1(5.19E-4)$^{\dag}$ & 1.074E-1(6.95E-6) & 1.077E-1(3.30E-4)  \\ \cline{2-8}
                         & 8   & \bb{1.993E-1(8.34E-3)} & 2.052E-1(4.98E-3) & 2.009E-1(4.97E-3) & 2.460E-1(1.11E-4)$^{\dag}$ & 2.031E-1(2.07E-3) & 2.011E-1(8.72E-4)  \\ \cline{2-8}
                         & 10  & \bb{2.104E-1(2.27E-4)} & 2.310E-1(2.52E-2)$^{\dag}$ & 2.151E-1(2.72E-3)$^{\dag}$ & 2.655E-1(7.16E-3)$^{\dag}$ & 2.154E-1(5.21E-3)$^{\dag}$ & 2.163E-1(3.30E-3)$^{\dag}$ \\ \cline{2-8}
                         & 15  & \bb{3.463E-1(4.76E-3)} & 3.686E-1(1.41E-2)$^{\dag}$ & 3.989E-1(8.25E-3)$^{\dag}$ & 3.688E-1(2.49E-2)$^{\dag}$ & 3.680E-1(8.14E-2)$^{\dag}$ & 3.909E-1(5.29E-2)$^{\dag}$ \\ \hline\hline
\multirow{5}{*}{C3-DTLZ4}& 3   & \bb{4.789E-1(2.00E-6)} & 4.838E-1(1.03E-4)$^{\dag}$ & 4.841E-1(4.21E-3)$^{\dag}$ & 4.848E-1(2.57E-4)$^{\dag}$ & 4.824E-1(3.57E-4)$^{\dag}$ & 4.813E-1(8.11E-4)$^{\dag}$ \\ \cline{2-8}
                         & 5   & \bb{4.170E-1(5.51E-4)} & 4.358E-1(5.65E-3)$^{\dag}$ & 4.484E-1(4.89E-3)$^{\dag}$ & 4.249E-1(5.17E-3)$^{\dag}$ & 4.430E-1(5.07E-3)$^{\dag}$ & 4.389E-1(1.36E-2)$^{\dag}$ \\ \cline{2-8}
                         & 8   & 5.049E-1(4.77E-4) & \bb{5.020E-1(5.33E-4)} & 5.268E-1(7.46E-3)$^{\dag}$ & 6.481E-1(1.35E-4)$^{\dag}$ & 5.234E-1(6.96E-3)$^{\dag}$ & 5.236E-1(3.33E-4)$^{\dag}$ \\ \cline{2-8}
                         & 10  & 5.604E-1(3.19E-3) & \bb{5.571E-1(5.34E-3)} & 5.651E-1(1.18E-3) & 5.735E-1(4.11E-3)$^{\dag}$ & 5.643E-1(2.22E-2) & 5.645E-1(8.09E-2) \\ \cline{2-8}
                         & 15  & 7.587E-1(5.23E-3)  & 7.627E-1(3.79E-2)$^{\dag}$  & 7.589E-1(4.40E-2)$^{\dag}$  & \bb{7.587E-1(3.78E-2)}  & 7.590E-1(8.28E-3)$^{\dag}$  & 7.589E-1(2.25E-2)$^{\dag}$ \\ \hline
\end{tabular}
}
\begin{tablenotes}
\item[1] $^{\dag}$ denotes the performance of C-TAEA is significantly better than the other peers according to the Wilcoxon's rank sum test at a 0.05 significance level; $^{\ddag}$ denotes the corresponding algorithm significantly outperforms C-TAEA.
\end{tablenotes} 
\end{table*}

\begin{table*}[htbp]
\centering 
\caption{Comparison results on HV metric (median and IQR) for C-TAEA and the other peer algorithms on C-DTLZ Benchmark Suite}
\label{tab:CDTLZ-HV}
\resizebox{\textwidth}{!}{ 
\begin{tabular}{c|c|c|c|c|c|c|c}
\hline
                         & $m$ & C-TAEA            & C-NSGA-III        & C-MOEA/D          & C-MOEA/DD         & I-DBEA           & CMOEA             \\ \hline
\multirow{5}{*}{C1-DTLZ1} & 3   & 1.3042(1.01E-3) & 1.3020(1.89E-3)$^{\dag}$ & \bb{1.3043(5.43E-4)} & 1.3043(1.07E-3) & 1.3033(2.42E-4) & 1.3039(1.60E-3) \\ \cline{2-8}
                          & 8   & 2.1435(3.02E-3) & 2.1431(6.59E-4) & 2.1436(1.00E-6) & \bb{2.1436(8.00E-6)} & 2.1436(6.01E-6) & 2.1436(1.35E-6)  \\ \cline{2-8}
                         & 10  & 2.5937(2.01E-6) & \bb{2.5940(3.52E-4)} & 2.5937(1.02E-6) & 2.5937(5.11E-6) & 2.5937(1.03E-6) & 2.5937(2.03E-6) \\ \cline{2-8}
                         & 15  & \bb{4.1022(3.83E-2)} & 4.0812(8.86E-2) & 4.0072(7.77E-2)$^{\dag}$ & 4.0277(9.15E-2) & 4.0911(7.93E-2) & 4.0288(3.35E-2)  \\ \hline\hline
\multirow{5}{*}{C1-DTLZ3} & 3  & \bb{0.7351(4.00E-2)} & 0.0000(0.00E+0)$^{\dag}$ & 0.0000(0.00E+0)$^{\dag}$ & 0.0000(0.00E+0)$^{\dag}$ & 0.0000(0.00E+0)$^{\dag}$ & 0.0000(0.00E+0)$^{\dag}$ \\ \cline{2-8}
                         & 5   & \bb{0.1761(1.76E-1)} & 0.0000(0.00E+0)$^{\dag}$ & 0.0000(0.00E+0)$^{\dag}$ & 0.0000(0.00E+0)$^{\dag}$ & 0.0000(0.00E+0)$^{\dag}$ & 0.0000(0.00E+0)$^{\dag}$ \\ \cline{2-8}
                         & 8   & \bb{1.5943(8.54E-2)} & 0.0000(0.00E+0)$^{\dag}$ & 0.0000(0.00E+0)$^{\dag}$ & 0.0000(0.00E+0)$^{\dag}$ & 0.0000(0.00E+0)$^{\dag}$ & 0.0000(0.00E+0)$^{\dag}$ \\ \cline{2-8}
                         & 10  & \bb{2.5132(6.34E-1)} & 0.0000(0.00E+0)$^{\dag}$ & 0.0000(0.00E+0)$^{\dag}$ & 0.0000(0.00E+0)$^{\dag}$ & 0.0000(0.00E+0)$^{\dag}$ & 0.0000(0.00E+0)$^{\dag}$ \\ \cline{2-8}
                         & 15  & \bb{2.1825(3.86E-2)} & 0.0000(0.00E+0)$^{\dag}$ & 0.0000(0.00E+0)$^{\dag}$ & 0.0000(0.00E+0)$^{\dag}$ & 0.0000(0.00E+0)$^{\dag}$ & 0.0000(0.00E+0)$^{\dag}$ \\ \hline\hline
\multirow{5}{*}{C2-DTLZ2} & 3  & \bb{0.4130(2.81E-4)} & 0.1225(1.00E-5)$^{\dag}$ & 0.1225(1.11E-1)$^{\dag}$ & 0.0000(1.21E-1)$^{\dag}$ & 0.1225(3.50E-5)$^{\dag}$ & 0.1225(6.50E-5)$^{\dag}$ \\ \cline{2-8}
                         & 5   & \bb{0.8607(2.50E-1)} & 0.1482(3.00E-6)$^{\dag}$ & 0.6117(4.63E-1)$^{\dag}$ & 0.1482(7.30E-3)$^{\dag}$ & 0.1482(6.13E-2)$^{\dag}$ & 0.1482(5.21E-4)$^{\dag}$ \\ \cline{2-8}
                         & 8   & \bb{1.1426(6.57E-3)} & 0.1973(2.90E-5)$^{\dag}$ & 0.1973(3.12E-6)$^{\dag}$ & 0.1949(7.21E-3)$^{\dag}$ & 0.3764(1.79E-1)$^{\dag}$ & 0.1973(2.32E-6)$^{\dag}$ \\ \cline{2-8}
                         & 10  & \bb{1.5937(2.70E-5)} & 0.2387(1.20E-5)$^{\dag}$ & 0.2387(2.03E-1)$^{\dag}$ & 0.2358(7.21E-2)$^{\dag}$ & 0.2386(2.17E-1)$^{\dag}$ & 0.2387(3.71E-6)$^{\dag}$ \\ \cline{2-8}
                         & 15  & \bb{2.4033(6.90E-2)} & 0.3840(5.90E-3)$^{\dag}$ & 0.3843(7.63E-2)$^{\dag}$ & 0.3797(9.26E-2)$^{\dag}$ & 0.3845(5.40E-2)$^{\dag}$ & 0.3843(4.26E-2)$^{\dag}$ \\ \hline\hline
\multirow{5}{*}{C3-DTLZ1} & 3  & \bb{1.1515(5.15E-4)} & 1.1499(1.46E-2) & 1.1253(2.37E-2)$^{\dag}$ & 1.1086(5.30E-4)$^{\dag}$ & 1.1310(3.56E-2) & 1.1427(6.03E-6) \\ \cline{2-8}
                         & 5   & \bb{1.5781(2.15E-4)} & 1.5736(1.28E-2) & 1.5776(7.27E-4) & 1.5656(2.20E-5)$^{\dag}$ & 1.5780(1.04E-4) & 1.5779(7.30E-5)  \\ \cline{2-8}
                         & 8   & 2.1386(5.66E-4) & 2.1332(3.07E-3) & \bb{2.1386(3.14E-4)} & 2.1367(4.00E-6) & 2.1385(5.80E-5) & 2.1386(9.00E-6)  \\ \cline{2-8}
                         & 10  & 2.5929(2.20E-5) & 2.5890(3.95E-3) & \bb{2.5929(1.50E-5)} & 2.5926(1.72E-2) & 2.5929(7.36E-2) & 2.5929(2.11E-2) \\ \cline{2-8}
                         & 15  & 4.1422(3.68E-1) & \bb{4.1769(5.67E-1)} & 4.1631(4.29E-2) & 4.1701(7.82E-2) & 4.1202(5.30E-2)& 4.1663(8.62E-2) \\ \hline\hline
\multirow{5}{*}{C3-DTLZ4}& 3   &8.4280(1.23E-2) &  \bb{8.4280(1.16E-3)} & 8.4165(6.70E-3)$^{\dag}$ & 8.4161(1.35E-2)$^{\dag}$ & 8.4150(9.29E-2)$^{\dag}$ & 8.4166(8.02E-2)$^{\dag}$ \\ \cline{2-8}
                         & 5   & \bb{49.5453(2.20E-3)} & 49.5451(7.20E-2) & 49.5330(5.80E-3)$^{\dag}$ & 49.5327(6.90E-3)$^{\dag}$ & 49.5346(1.67E-2)$^{\dag}$ & 49.5257(3.93E-2)$^{\dag}$ \\ \cline{2-8}
                         & 8   & 546.4971(4.56E-2) & \bb{546.4971(1.10E-3)} & 546.4951(4.20E-3) & 546.4943(2.29E-2) & 546.4933(3.73E-2) & 546.4942(4.21E-2) \\ \cline{2-8}
                         & 10  & 2654.4042(9.19E-2) & \bb{2654.4042(7.84E-2)} & 2654.4042(5.37E-2) & 2654.4041(1.98E-2) & 2654.4042(3.24E-2) & 2654.4042(3.15E-2) \\ \cline{2-8}
                         & 15  & \bb{136803.0202(4.13E-2)}  & 136802.2201(3.70E-2)$^{\dag}$  & 136802.2302(5.26E-2)$^{\dag}$  & 136802.1233(9.10E-2)$^{\dag}$  & 136802.1921(9.80E-0)$^{\dag}$  & 136802.0201(6.23E-1)$^{\dag}$ \\ \hline
\end{tabular}
}
\begin{tablenotes}
\item[1] $^{\dag}$ denotes the performance of C-TAEA is significantly better than the other peers according to the Wilcoxon's rank sum test at a 0.05 significance level; $^{\ddag}$ denotes the corresponding algorithm significantly outperforms C-TAEA.
\end{tablenotes} 
\end{table*}

The comparison results of IGD and HV values are given in~\pref{tab:CDTLZ-IGD} and~\pref{tab:CDTLZ-HV} respectively. Generally speaking, our proposed C-TAEA produces superior IGD and HV values on most test instances.

\begin{figure}[htbp]
	\centering
    \includegraphics[width=.7\linewidth]{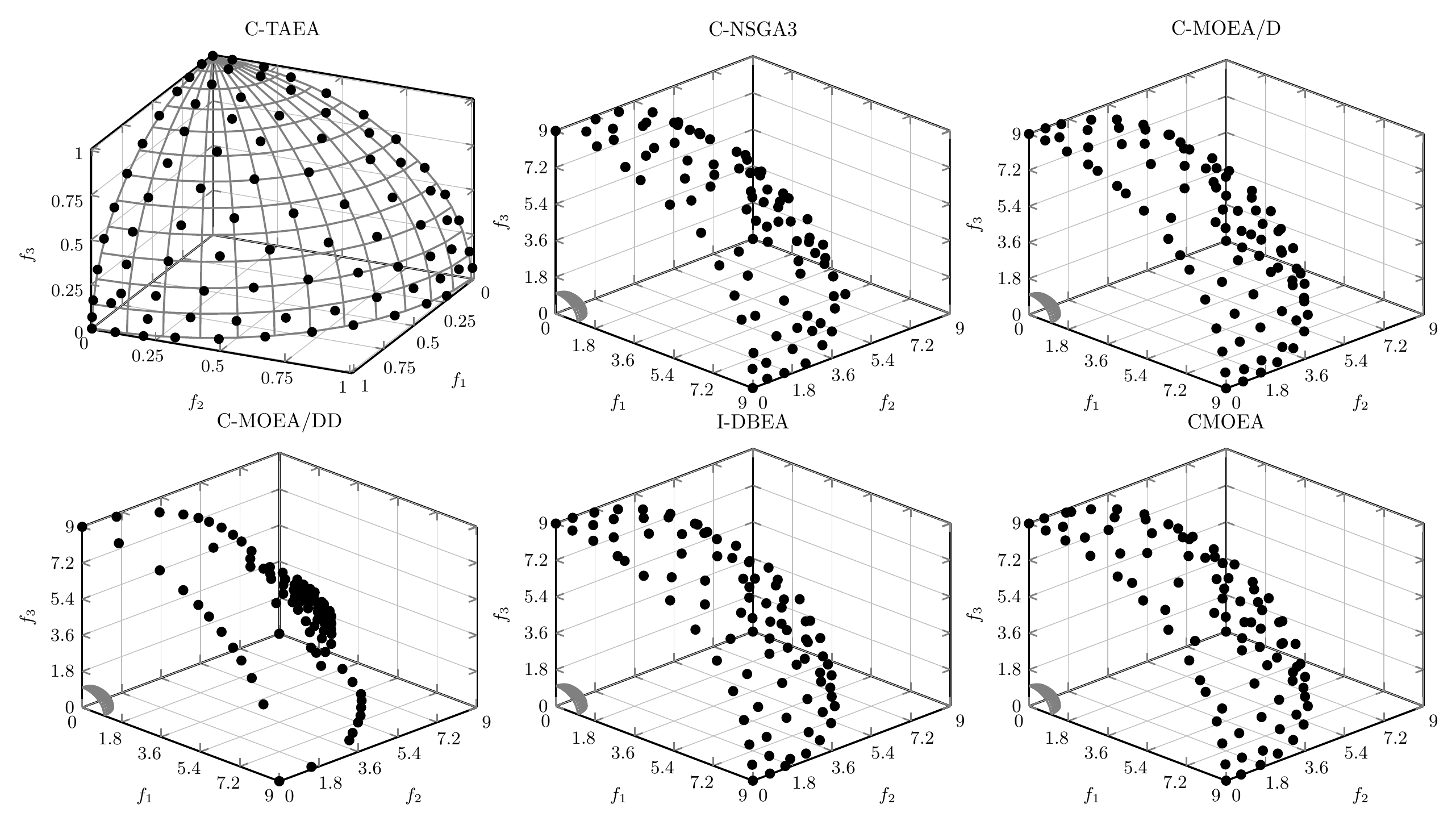}
    \caption{Scatter plots of the population obtained by C-TAEA and the peer algorithms on C1-DTLZ3 (median IGD value).}
    \label{fig:C1DTLZ3}
\end{figure}

Let us first look at the Type-1 constrained problem. Although the feasible region of C1-DTLZ1 is only a narrow region above the PF, it actually does not pose any difficulty to all algorithms. In particular, all algorithms, especially those purely feasibility-driven ones, just simply push solutions toward the feasible boundary. As for C1-DTLZ3, C-TAEA shows the best performance on all 3- to 15-objective problem instances. In particular, it obtains around 50 times smaller IGD values than the other peer algorithms on average; only C-TAEA obtains effective HV values while the HV values obtained by the other peer algorithms are all 0, which means that the obtained non-dominated solutions are all dominated by $\mathbf{z}^r$. As shown in Fig. 2 of the supplementary document, C1-DTLZ3 places an infeasible barrier in the attainable objective space, which obstructs the population for converging to the true PF. As discussed in~\pref{sec:motivations}, due to their feasibility-driven selection strategy, the other peer algorithms cannot provide any further selection pressure to push the population forward when it approaches the outer boundary of this infeasible barrier, as shown in~\pref{fig:C1DTLZ3}\footnote{We only show the 3-objective scatter plots in this paper, while the high-dimensional plots, which are not as intuitive as the 3-objective scenarios, are put in the supplementary document.}. In contrast, since the selection mechanism of the DA does not take the feasibility information into account, it can constantly push the solutions of the DA toward the PF without considering the existence of this infeasible barrier. In the meanwhile, the CA can at the end overcome this infeasible barrier via the restricted mating selection between the CA and DA. We also notice that C-TAEA cannot push solutions to fully converge on the PF in high-dimensional cases as shown in Fig. 17 to 20 of the supplementary document. This is because the size of the infeasible barrier increases with the dimensionality. It makes C1-DTLZ3 even more difficult in a many-objective scenario. Nevertheless, the solutions obtained by C-TAEA are much closer to the PF than the other peer algorithms.

\begin{figure}[htbp]
	\centering
    \includegraphics[width=.7\linewidth]{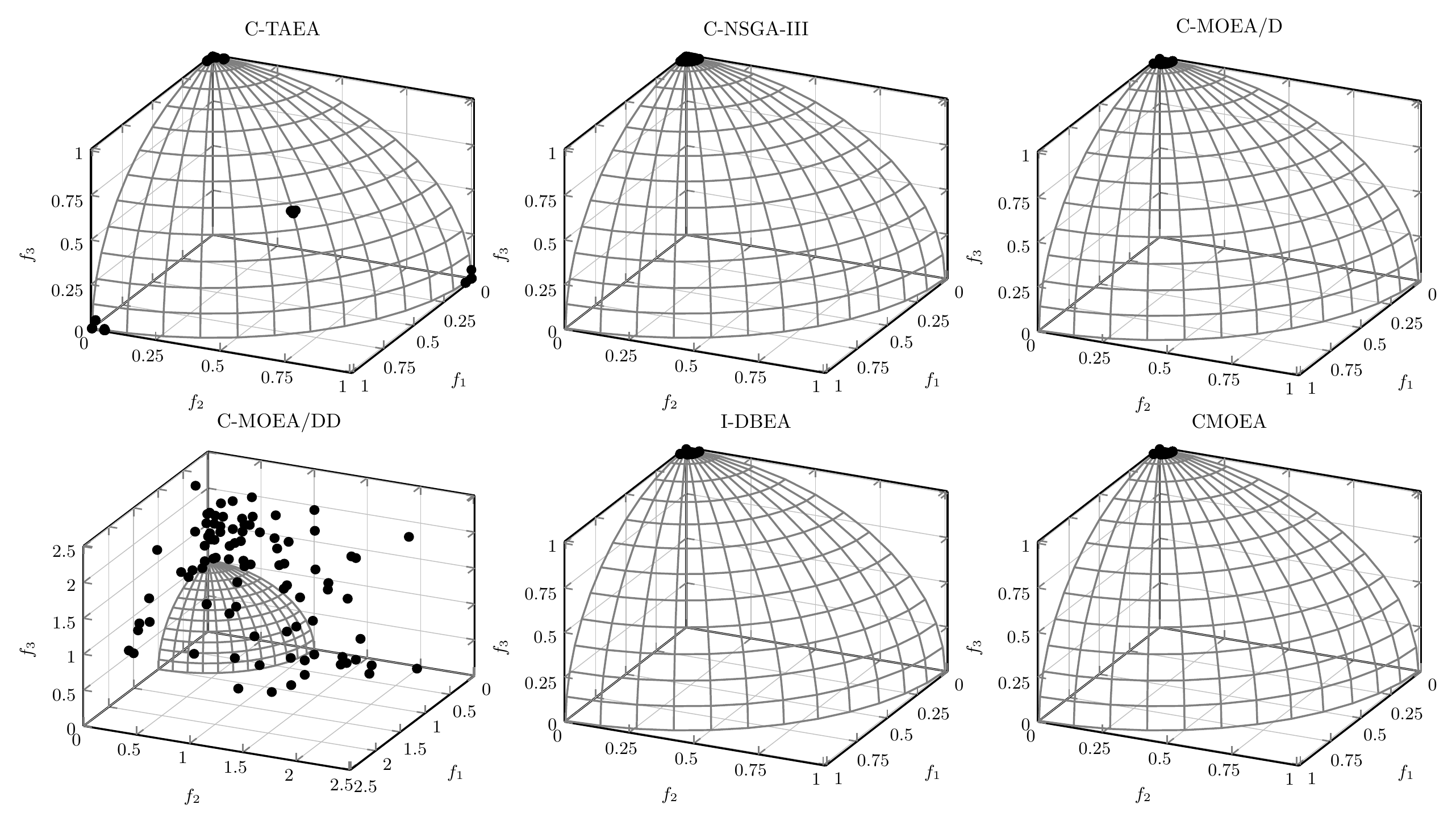}
    \caption{Scatter plots of the population obtained by C-TAEA and the peer algorithms on C2-DTLZ2 (median IGD value).}
    \label{fig:C2DTLZ2}
\end{figure}

\begin{figure}[htbp]
	\centering
    \includegraphics[width=.7\linewidth]{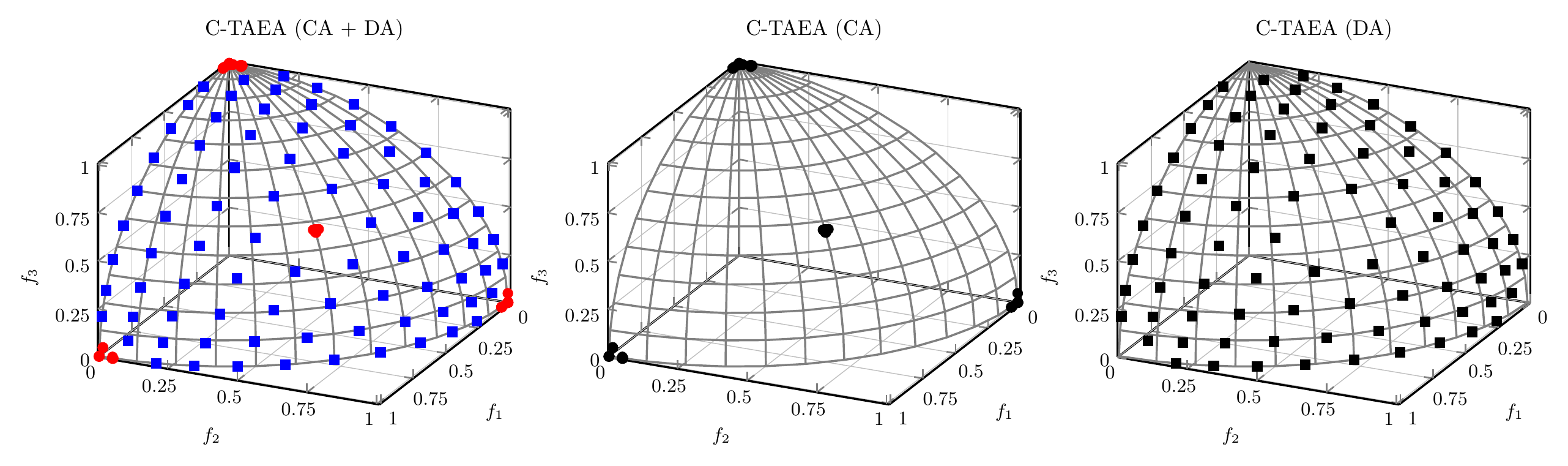}
    \caption{Comparison of the solutions finally obtained in CA and DA on C2-DTLZ2 (median IGD value).}
    \label{fig:CA+DA}
\end{figure}

The Type-2 constrained problem, i.e., C2-DTLZ2, spreads several feasible regions on disparate parts of the PF. All algorithms do not have any difficulty in finding at least one feasible PF segment, whereas only C-TAEA can find all disparately distributed small feasible PF segments as shown in~\pref{fig:C2DTLZ2}. The reason that leads to this phenomenon is similar to C1-DTLZ3. Specifically, each feasible region is small when setting a small $r$ in C2-DTLZ2, thus different feasible regions are separated by large infeasible barriers. In this case, if an algorithm finds one of the feasible PF segments, it hardly has a sufficient selection pressure to jump over this local feasible PF segment. However, due to the existence of the DA in C-TAEA, it complements the coverage of the CA. As shown in~\pref{fig:CA+DA}, solutions in the CA and DA perfectly complements each other in terms of the coverage over the PF. As a result, the DA helps drive the CA explore new feasible regions. 

As for the Type-3 constrained problems, i.e., C3-DTLZ1 and C3-DTLZ4, the original PF of the baseline problem becomes infeasible when considering the constraints while the new PF is formed by the feasible boundaries. In terms of the constraint handling, this type of problems does not provide too much difficulty. From the comparison results shown in~\pref{tab:CDTLZ-IGD} and~\pref{tab:CDTLZ-HV}, we find that all algorithms obtain comparable IGD and HV values on all C3-DTLZ1 and C3-DTLZ4 problem instances. In particular, C-TAEA is outperformed by C-MOEA/D on the 5-objective C3-DTLZ1 problem instance; and it is outperformed by C-NSGA-II on the 8- and 10-objective C3-DTLZ4 problem instances. In general, due to the advanced selection mechanisms of the CA and DA for balancing convergence and diversity, C-TAEA obtains better IGD and HV values on most cases.

\subsection{DC-DTLZ Benchmark Suite}
\label{sec:DCDTLZ}

\begin{table*}[htbp]
\centering
\scriptsize
\caption{Comparison results on IGD metric (median and IQR) for C-TAEA and the other peer algorithms on DC-DTLZ Benchmark Suite}
\label{tab:DCDTLZ-IGD}
\resizebox{\textwidth}{!}{
\begin{tabular}{c|c|c|c|c|c|c|c}
\hline
                         & $m$ & C-TAEA            & C-NSGA-III        & C-MOEA/D          & C-MOEA/DD         & I-DBEA           & CMOEA             \\ \hline
\multirow{5}{*}{DC1-DTLZ1}& 3   & \bb{5.638E-2(8.10E-5)} & 5.990E-2(1.59E-5)$^{\dag}$ & 1.835E-1(1.26E-1)$^{\dag}$ & 1.042E-1(2.03E-3)$^{\dag}$ & 5.843E-2(3.38E-3)$^{\dag}$ & 5.843E-2(3.65E-3)$^{\dag}$ \\ \cline{2-8} 
                          & 5   & \bb{7.301E-2(3.76E-3)} & 7.655E-2(1.41E-2)$^{\dag}$ & 7.327E-2(1.71E-4)$^{\dag}$ & 8.705E-2(3.75E-3)$^{\dag}$ & 7.344E-2(3.91E-4)$^{\dag}$ & 7.640E-2(3.38E-3)$^{\dag}$ \\ \cline{2-8} 
                          & 8   & \bb{1.086E-1(6.44E-4)} & 1.104E-1(9.78E-4)$^{\dag}$ & 1.414E-1(1.51E-2)$^{\dag}$ & 1.175E-1(7.30E-2)$^{\dag}$ & 1.290E-1(8.13E-2)$^{\dag}$ & 1.291E-1(5.20E-3)$^{\dag}$ \\ \cline{2-8} 
                          & 10  & \bb{1.189E-1(2.84E-3)} & 1.206E-1(3.34E-3)$^{\dag}$ & 1.524E-1(5.84E-3)$^{\dag}$ & 1.278E-1(4.24E-2)$^{\dag}$ & 1.545E-1(6.27E-2)$^{\dag}$ & 1.529E-1(8.16E-3)$^{\dag}$ \\ \cline{2-8} 
                          & 15  & \bb{1.753E-1(1.83E-2)} & 1.984E-1(4.11E-3)$^{\dag}$ & 2.017E-1(6.17E-2)$^{\dag}$ & 1.772E-1(5.25E-3)$^{\dag}$ & 2.070E-1(7.79E-2)$^{\dag}$ & 1.986E-1(2.06E-2)$^{\dag}$ \\ \hline\hline 
\multirow{5}{*}{DC1-DTLZ3}& 3   & \bb{1.466E-1(7.62E-4)} & 2.720E-1(1.31E-1)$^{\dag}$ & 1.349E-1(3.77E-1)$^{\dag}$ & 2.908E-1(1.18E-1)$^{\dag}$ & 5.140E-1(3.75E-1)$^{\dag}$ & 5.140E-1(3.77E-1)$^{\dag}$ \\ \cline{2-8} 
                          & 5   & \bb{2.083E-1(2.54E-3)} & 2.040E-1(1.01E-2)$^{\dag}$ & 3.947E-1(1.18E-4)$^{\dag}$ & 2.318E-1(7.15E-4)$^{\dag}$ & 3.948E-1(8.69E-4)$^{\dag}$ & 3.947E-1(2.47E-4)$^{\dag}$ \\ \cline{2-8} 
                          & 8   & \bb{3.405E-1(8.35E-5)} & 4.062E-1(3.03E-2)$^{\dag}$ & 4.330E-1(1.68E-3)$^{\dag}$ & 3.639E-1(5.28E-2)$^{\dag}$ & 4.344E-1(8.13E-3)$^{\dag}$ & 3.422E-1(4.01E-2)$^{\dag}$ \\ \cline{2-8} 
                          & 10  & \bb{3.886E-1(3.18E-3)} & 4.586E-1(4.89E-2)$^{\dag}$ & 4.596E-1(4.06E-3)$^{\dag}$ & 4.154E-1(9.14E-3)$^{\dag}$ & 4.456E-1(2.23E-3)$^{\dag}$ & 4.235E-1(5.23E-3)$^{\dag}$ \\ \cline{2-8} 
                          & 15  & \bb{8.009E-1(5.10E-3)} & 8.287E-1(6.23E-3)$^{\dag}$ & 8.456E-1(6.28E-2)$^{\dag}$ & 8.034E-1(5.80E-3)$^{\dag}$ & 8.150E-1(1.26E-2)$^{\dag}$ & 8.144E-1(7.20E-3)$^{\dag}$ \\ \hline\hline 
\multirow{5}{*}{DC2-DTLZ1}& 3   & \bb{2.199E-2(8.44E-3)} & 1.760E+1(3.28E-2)$^{\dag}$ & 2.328E+1(3.38E-2)$^{\dag}$ & 2.744E+1(5.08E-2)$^{\dag}$ & 2.167E+1(8.49E-2)$^{\dag}$ & 2.248E+1(7.27E-2)$^{\dag}$ \\ \cline{2-8} 
                          & 5   & \bb{5.371E-2(3.07E-2)} & 1.644E+1(8.56E-2)$^{\dag}$ & 1.308E+1(1.24E-2)$^{\dag}$ & 2.696E+1(9.03E-2)$^{\dag}$ & 1.684E+1(5.80E-2)$^{\dag}$ & 1.894E+1(5.53E-2)$^{\dag}$ \\ \cline{2-8} 
                          & 8   & \bb{9.937E-2(8.10E-2)} & 1.275E+1(3.69E-2)$^{\dag}$ & 2.033E+1(4.32E-2)$^{\dag}$ & 2.393E+1(3.55E-2)$^{\dag}$ & 1.774E+1(4.34E-3)$^{\dag}$ & 2.629E+1(7.50E-2)$^{\dag}$ \\ \cline{2-8} 
                          & 10  & \bb{1.048E-1(8.65E-3)} & 1.056E-1(8.09E-2)$^{\dag}$ & 1.489E+1(2.84E-2)$^{\dag}$ & 1.851E+1(1.76E-2)$^{\dag}$ & 1.058E-1(7.89E-2)$^{\dag}$ & 2.176E+1(5.82E-2)$^{\dag}$ \\ \cline{2-8} 
                          & 15  & \bb{2.308E-1(4.02E-2)} & 1.720E+1(6.54E-2)$^{\dag}$ & 1.918E+1(7.47E-2)$^{\dag}$ & 1.510E+1(4./7E-2)$^{\dag}$ & 1.881E+1(9.25E-2)$^{\dag}$ & 1.564E+1(6.04E-3)$^{\dag}$ \\ \hline\hline 
\multirow{5}{*}{DC2-DTLZ3}& 3   & \bb{5.498E-2(6.78E-2)} & 1.203E+2(3.71E-2)$^{\dag}$ & 5.668E+1(7.42E-2)$^{\dag}$ & 1.204E+2(1.18E-3)$^{\dag}$ & 1.202E+2(2.29E-2)$^{\dag}$ & 1.204E+2(5.93E-2)$^{\dag}$ \\ \cline{2-8} 
                          & 5   & \bb{1.667E-1(9.36E-3)} & 5.692E+1(5.38E-2)$^{\dag}$ & 5.657E+1(8.00E-2)$^{\dag}$ & 1.204E+2(5.72E-2)$^{\dag}$ & 1.204E+2(4.31E-2)$^{\dag}$ & 5.679E+1(3.80E-2)$^{\dag}$ \\ \cline{2-8} 
                          & 8   & \bb{5.674E+1(4.65E-2)} & 5.701E+1(6.01E-2)$^{\dag}$ & 1.206E+2(9.71E-3)$^{\dag}$ & 5.701E+1(9.00E-2)$^{\dag}$ & 5.688E+1(3.18E-2)$^{\dag}$ & 1.206E+2(4.85E-2)$^{\dag}$ \\ \cline{2-8} 
                          & 10  & \bb{3.836E-1(6.50E-2)} & 5.715E+1(7.56E-2)$^{\dag}$ & 5.700E+1(3.04E-2)$^{\dag}$ & 1.206E+2(2.76E-2)$^{\dag}$ & 5.701E+1(2.92E-2)$^{\dag}$ & 1.205E+2(4.35E-2)$^{\dag}$ \\ \cline{2-8} 
                          & 15  & \bb{7.959E-1(8.68E-2)} & 5.726E+1(6.86E-2)$^{\dag}$ & 1.208E+2(4.42E-2)$^{\dag}$ & 1.209E+2(6.18E-2)$^{\dag}$ & 5.734E+1(4.44E-2)$^{\dag}$ & 5.729E+1(7.27E-2)$^{\dag}$ \\ \hline\hline 
\multirow{5}{*}{DC3-DTLZ1}& 3   & \bb{5.034e-2(1.72E-4)} & 9.745E+0(5.64E-3)$^{\dag}$ & 9.746E+0(7.80E-3)$^{\dag}$ & 9.789E+0(8.76E-4)$^{\dag}$ & 9.745E+0(2.02E-3)$^{\dag}$ & 9.755E+0(1.29E-2)$^{\dag}$ \\ \cline{2-8} 
                          & 5   & \bb{8.554E-1(1.29E-3)} & 7.702E+0(2.60E-2)$^{\dag}$ & 8.165E+0(1.78E-1)$^{\dag}$ & 8.467E+0(1.21E-1)$^{\dag}$ & 1.847E+1(1.03E+1)$^{\dag}$ & 8.408E+0(1.71E-3)$^{\dag}$ \\ \cline{2-8} 
                          & 8   & \bb{1.250E-1(6.01E-1)} & 6.450E+0(2.30E+0)$^{\dag}$ & 9.729E+0(2.03E+0)$^{\dag}$ & 6.988E+0(3.74E-3)$^{\dag}$ & 8.409E+0(1.30E-2)$^{\dag}$ & 5.938E+0(2.83E+0)$^{\dag}$ \\ \cline{2-8} 
                          & 10  & \bb{2.332E-1(5.29E-3)} & 5.598E+0(8.71E-2)$^{\dag}$ & 2.120E+1(7.29E-3)$^{\dag}$ & 6.004E+0(8.26E-3)$^{\dag}$ & 8.432E+0(5./9E-2)$^{\dag}$ & 7.166E+0(1.93E-3)$^{\dag}$ \\ \cline{2-8} 
                          & 15  & \bb{1.837E-1(3.43E-5)} & 5.431E+0(4.38E-1)$^{\dag}$ & 2.567E+1(1.10E+1)$^{\dag}$ & 2.346E-1(7.51E+0)$^{\dag}$ & 7.204E+0(1.76E+1)$^{\dag}$ & 2.584E+1(1.66E+1)$^{\dag}$ \\ \hline\hline
\multirow{5}{*}{DC3-DTLZ3}& 3   & \bb{1.250E-1(8.04E-4)} & 3.334E+1(7.20E-2)$^{\dag}$ & 3.335E+1(6.20E-2)$^{\dag}$ & 3.337E+1(2.54E-2)$^{\dag}$ & 7.335E+1(8.46E-2)$^{\dag}$ & 7.335E+1(4.52E-2)$^{\dag}$ \\ \cline{2-8} 
                          & 5   & \bb{2.219E-1(9.16E-3)} & 3.349E+1(5.57E-3)$^{\dag}$ & 3.340E+1(3.75E-3)$^{\dag}$ & 3.341E+1(4.86E-4)$^{\dag}$ & 3.340E+1(7.59E-1)$^{\dag}$ & 3.339E+1(2.28E-2)$^{\dag}$ \\ \cline{2-8} 
                          & 8   & \bb{3.429E-1(8.37E-2)} & 3.360E+1(3.52E-3)$^{\dag}$ & 3.350E+1(1.88E-2)$^{\dag}$ & 3.343E+1(5.02E-3)$^{\dag}$ & 3.369E+1(3.39E-3)$^{\dag}$ & 3.359E+1(7.59E-3)$^{\dag}$ \\ \cline{2-8} 
                          & 10  & \bb{3.835E-1(1.16E-3)} & 3.362E+1(9.10E-2)$^{\dag}$ & 7.377E+1(9.92E-3)$^{\dag}$ & 7.346E+1(8.57E-3)$^{\dag}$ & 7.376E+1(7.36E-2)$^{\dag}$ & 7.377E+1(4.91E-2)$^{\dag}$ \\ \cline{2-8} 
                          & 15  & \bb{7.872E-1(2.33E-2)} & 7.411E+1(3.62E-3)$^{\dag}$ & 1.541E+2(8.61E-3)$^{\dag}$ & 7.407E+1(9.35E-2)$^{\dag}$ & 7.416E+1(4.29E-2)$^{\dag}$ & 7.407E+1(5.49E-2)$^{\dag}$ \\ \hline 
\end{tabular}
}
\begin{tablenotes}
\item[1] $^{\dag}$ denotes the performance of C-TAEA is significantly better than the other peers according to the Wilcoxon's rank sum test at a 0.05 significance level; $^{\ddag}$ denotes the corresponding algorithm significantly outperforms C-TAEA.
\end{tablenotes} 
\end{table*}

\begin{table*}[htbp]
\centering 
\caption{Comparison results on HV metric (median and IQR) for C-TAEA and the other peer algorithms on DC-DTLZ Benchmark Suite}
\label{tab:DCDTLZ-HV}
\resizebox{\textwidth}{!}{ 
\begin{tabular}{c|c|c|c|c|c|c|c}
\hline
                         & $m$ & C-TAEA            & C-NSGA-III        & C-MOEA/D          & C-MOEA/DD         & I-DBEA           & CMOEA             \\ \hline
\multirow{5}{*}{DC1-DTLZ1}& 3   & \bb{1.2006(1.70E-2)} & 1.1982(9.96E-2)$^{\dag}$ & 0.9631(2.81E-2)$^{\dag}$ & 1.1845(3.05E-2)$^{\dag}$ & 1.1883(9.25E-2)$^{\dag}$ & 1.1883(8.40E-3)$^{\dag}$ \\ \cline{2-8} 
                          & 5   & \bb{1.4783(3.27E-2)} & 1.4725(3.36E-2)$^{\dag}$ & 1.4783(5.05E-2)$^{\dag}$ & 1.4762(8.46E-2)$^{\dag}$ & 1.4782(7.29E-2)$^{\dag}$ & 1.4779(3.13E-2)$^{\dag}$ \\ \cline{2-8} 
                          & 8  & \bb{1.9682(1.24E-2)}  & 1.9347(8.57E-2)$^{\dag}$ & 1.9660(8.95E-2)$^{\dag}$ & 1.9655(5.82E-2)$^{\dag}$ & 1.9678(5.45E-2)$^{\dag}$ & 1.9676(8.14E-2)$^{\dag}$ \\ \cline{2-8} 
                          & 10  & \bb{2.3890(3.67E-3)} & 2.3113(4.34E-2)$^{\dag}$ & 2.3792(3.64E-2)$^{\dag}$ & 2.3801(4.30E-3)$^{\dag}$ & 2.3797(7.50E-2)$^{\dag}$ & 2.3810(8.70E-3)$^{\dag}$ \\ \cline{2-8} 
                          & 15  & \bb{4.1012(1.73E-2)} & 4.0031(2.76E-2)$^{\dag}$ & 4.0122(1.78E-2)$^{\dag}$ & 4.0823(7.88E-2)$^{\dag}$ & 4.0911(5.84E-2)$^{\dag}$ & 4.0532(4.03E-2)$^{\dag}$ \\ \hline\hline 
\multirow{5}{*}{DC1-DTLZ3}& 3   & \bb{0.6339(6.51E-2)} & 0.5088(7.54E-2)$^{\dag}$ & 0.6694(3.99E-2)$^{\dag}$ & 0.5386(9.32E-2)$^{\dag}$ & 0.4545(6.00E-3)$^{\dag}$ & 0.4545(6.76E-2)$^{\dag}$ \\ \cline{2-8} 
                          & 5   & \bb{1.2656(3.68E-2)} & 1.2582(7.39E-2)$^{\dag}$ & 1.1463(1.20E-3)$^{\dag}$ & 1.1712(2.26E-2)$^{\dag}$ & 1.1467(5.86E-2)$^{\dag}$ & 1.1462(9.40E-3)$^{\dag}$ \\ \cline{2-8} 
                          & 8   & \bb{1.9829(5.39E-2)} & 1.8461(7.95E-2)$^{\dag}$ & 1.9301(5.70E-2)$^{\dag}$ & 1.9640(4.34E-2)$^{\dag}$ & 1.9284(3.78E-2)$^{\dag}$ & 1.9793(4.67E-2)$^{\dag}$ \\ \cline{2-8} 
                          & 10  & \bb{2.5181(6.01E-2)} & 2.3880(9.70E-3)$^{\dag}$ & 2.4903(9.02E-2)$^{\dag}$ & 2.5083(3.17E-2)$^{\dag}$ & 2.4902(4.92E-2)$^{\dag}$ & 2.4986(6.52E-2)$^{\dag}$ \\ \cline{2-8} 
                          & 15  & \bb{4.1700(7.56E-2)} & 4.0321(3.01E-2)$^{\dag}$ & 4.0422(2.80E-2)$^{\dag}$ & 4.0282(2.86E-2)$^{\dag}$ & 3.9123(4.41E-2)$^{\dag}$ & 4.1028(8.65E-2)$^{\dag}$ \\ \hline\hline 
\multirow{5}{*}{DC2-DTLZ1}& 3   & \bb{1.1610(5.15E-4)} & 0.0000(0.00E+0)$^{\dag}$ & 0.0000(0.00E+0)$^{\dag}$ & 0.0000(0.00E+0)$^{\dag}$ & 0.0000(0.00E+0)$^{\dag}$ & 0.0000(0.00E+0)$^{\dag}$ \\ \cline{2-8} 
                          & 5   & \bb{1.5781(2.15E-4)} & 0.0000(0.00E+0)$^{\dag}$ & 0.0000(0.00E+0)$^{\dag}$ & 0.0000(0.00E+0)$^{\dag}$ & 0.0000(0.00E+0)$^{\dag}$ & 0.0000(0.00E+0)$^{\dag}$ \\ \cline{2-8} 
                          & 8   & \bb{2.1386(5.66E-4)} & 0.0000(0.00E+0)$^{\dag}$ & 0.0000(0.00E+0)$^{\dag}$ & 0.0000(0.00E+0)$^{\dag}$ & 0.0000(0.00E+0)$^{\dag}$ & 0.0000(0.00E+0)$^{\dag}$ \\ \cline{2-8} 
                          & 10  & \bb{2.5929(2.20E-5)} & 0.0000(0.00E+0)$^{\dag}$ & 0.0000(0.00E+0)$^{\dag}$ & 0.0000(0.00E+0)$^{\dag}$ & 0.0000(0.00E+0)$^{\dag}$ & 0.0000(0.00E+0)$^{\dag}$ \\ \cline{2-8} 
                          & 15  & \bb{4.1422(3.68E-1)} & 0.0000(0.00E+0)$^{\dag}$ & 0.0000(0.00E+0)$^{\dag}$ & 0.0000(0.00E+0)$^{\dag}$ & 0.0000(0.00E+0)$^{\dag}$ & 0.0000(0.00E+0)$^{\dag}$ \\ \hline\hline 
\multirow{5}{*}{DC2-DTLZ3}& 3   & \bb{0.7377(3.68E-2)} & 0.0000(0.00E+0)$^{\dag}$ & 0.0000(0.00E+0)$^{\dag}$ & 0.0000(0.00E+0)$^{\dag}$ & 0.0000(0.00E+0)$^{\dag}$ & 0.0000(0.00E+0)$^{\dag}$ \\ \cline{2-8} 
                          & 5   & \bb{1.3087(7.23E-2)} & 0.0000(0.00E+0)$^{\dag}$ & 0.0000(0.00E+0)$^{\dag}$ & 0.0000(0.00E+0)$^{\dag}$ & 0.0000(0.00E+0)$^{\dag}$ & 0.0000(0.00E+0)$^{\dag}$ \\ \cline{2-8} 
                          & 8   & \bb{2.0013(5.55E-2)} & 0.0000(0.00E+0)$^{\dag}$ & 0.0000(0.00E+0)$^{\dag}$ & 0.0000(0.00E+0)$^{\dag}$ & 0.0000(0.00E+0)$^{\dag}$ & 0.0000(0.00E+0)$^{\dag}$ \\ \cline{2-8} 
                          & 10  & \bb{2.5101(3.50E-2)} & 0.0000(0.00E+0)$^{\dag}$ & 0.0000(0.00E+0)$^{\dag}$ & 0.0000(0.00E+0)$^{\dag}$ & 0.0000(0.00E+0)$^{\dag}$ & 0.0000(0.00E+0)$^{\dag}$ \\ \cline{2-8} 
                          & 15  & \bb{4.0832(9.14E-2)} & 0.0000(0.00E+0)$^{\dag}$ & 0.0000(0.00E+0)$^{\dag}$ & 0.0000(0.00E+0)$^{\dag}$ & 0.0000(0.00E+0)$^{\dag}$ & 0.0000(0.00E+0)$^{\dag}$ \\ \hline\hline 
\multirow{5}{*}{DC3-DTLZ1}& 3   & \bb{1.2134(1.10E-5)} & 0.0000(0.00E+0)$^{\dag}$ & 0.0000(0.00E+0)$^{\dag}$ & 0.0000(0.00E+0)$^{\dag}$ & 0.0000(0.00E+0)$^{\dag}$ & 0.0000(0.00E+0)$^{\dag}$ \\ \cline{2-8} 
                          & 5   & \bb{1.4751(2.53E-3)} & 0.0000(0.00E+0)$^{\dag}$ & 0.0000(0.00E+0)$^{\dag}$ & 0.0000(0.00E+0)$^{\dag}$ & 0.0000(0.00E+0)$^{\dag}$ & 0.0000(0.00E+0)$^{\dag}$ \\ \cline{2-8} 
                          & 8   & \bb{1.9429(1.94E+0)} & 0.0000(0.00E+0)$^{\dag}$ & 0.0000(0.00E+0)$^{\dag}$ & 0.0000(0.00E+0)$^{\dag}$ & 0.0000(0.00E+0)$^{\dag}$ & 0.0000(0.00E+0)$^{\dag}$ \\ \cline{2-8} 
                          & 10  & \bb{2.3933(1.05E-1)} & 0.0000(0.00E+0)$^{\dag}$ & 0.0000(0.00E+0)$^{\dag}$ & 0.0000(0.00E+0)$^{\dag}$ & 0.0000(0.00E+0)$^{\dag}$ & 0.0000(0.00E+0)$^{\dag}$ \\ \cline{2-8} 
                          & 15  & \bb{4.0012(4.32E-2)} & 0.0000(0.00E+0)$^{\dag}$ & 0.0000(0.00E+0)$^{\dag}$ & 0.0000(0.00E+0)$^{\dag}$ & 0.0000(0.00E+0)$^{\dag}$ & 0.0000(0.00E+0)$^{\dag}$ \\ \hline\hline
\multirow{5}{*}{DC3-DTLZ3}& 3   & \bb{0.6298(4.74E-2)} & 0.0000(0.00E+0)$^{\dag}$ & 0.0000(0.00E+0)$^{\dag}$ & 0.0000(0.00E+0)$^{\dag}$ & 0.0000(0.00E+0)$^{\dag}$ & 0.0000(0.00E+0)$^{\dag}$ \\ \cline{2-8} 
                          & 5   & \bb{1.1880(2.35E-2)} & 0.0000(0.00E+0)$^{\dag}$ & 0.0000(0.00E+0)$^{\dag}$ & 0.0000(0.00E+0)$^{\dag}$ & 0.0000(0.00E+0)$^{\dag}$ & 0.0000(0.00E+0)$^{\dag}$ \\ \cline{2-8} 
                          & 8   & \bb{1.7614(9.28E-2)} & 0.0000(0.00E+0)$^{\dag}$ & 0.0000(0.00E+0)$^{\dag}$ & 0.0000(0.00E+0)$^{\dag}$ & 0.0000(0.00E+0)$^{\dag}$ & 0.0000(0.00E+0)$^{\dag}$ \\ \cline{2-8} 
                          & 10  & \bb{2.3748(6.13E-2)} & 0.0000(0.00E+0)$^{\dag}$ & 0.0000(0.00E+0)$^{\dag}$ & 0.0000(0.00E+0)$^{\dag}$ & 0.0000(0.00E+0)$^{\dag}$ & 0.0000(0.00E+0)$^{\dag}$ \\ \cline{2-8} 
                          & 15  & \bb{4.1326(1.28E-2)} & 0.0000(0.00E+0)$^{\dag}$ & 0.0000(0.00E+0)$^{\dag}$ & 0.0000(0.00E+0)$^{\dag}$ & 0.0000(0.00E+0)$^{\dag}$ & 0.0000(0.00E+0)$^{\dag}$ \\ \hline 
\end{tabular}
}
\begin{tablenotes}
\item[1] $^{\dag}$ denotes the performance of C-TAEA is significantly better than the other peers according to the Wilcoxon's rank sum test at a 0.05 significance level; $^{\ddag}$ denotes the corresponding algorithm significantly outperforms C-TAEA.
\end{tablenotes} 
\end{table*}

The comparison results of IGD and HV values on the DC-DTLZ benchmark suite are given in~\pref{tab:DCDTLZ-IGD} and~\pref{tab:DCDTLZ-HV} respectively. From these results, it is obvious to see the overwhelmingly superior performance of C-TAEA over the other peer algorithms, given the observation that C-TAEA obtains the best IGD and HV values in all comparisons. The following paragraphs try to decipher the potential reasons that lead to the ineffectiveness of the other peer algorithms.

\begin{figure}[htbp]
	\centering
    \includegraphics[width=.7\linewidth]{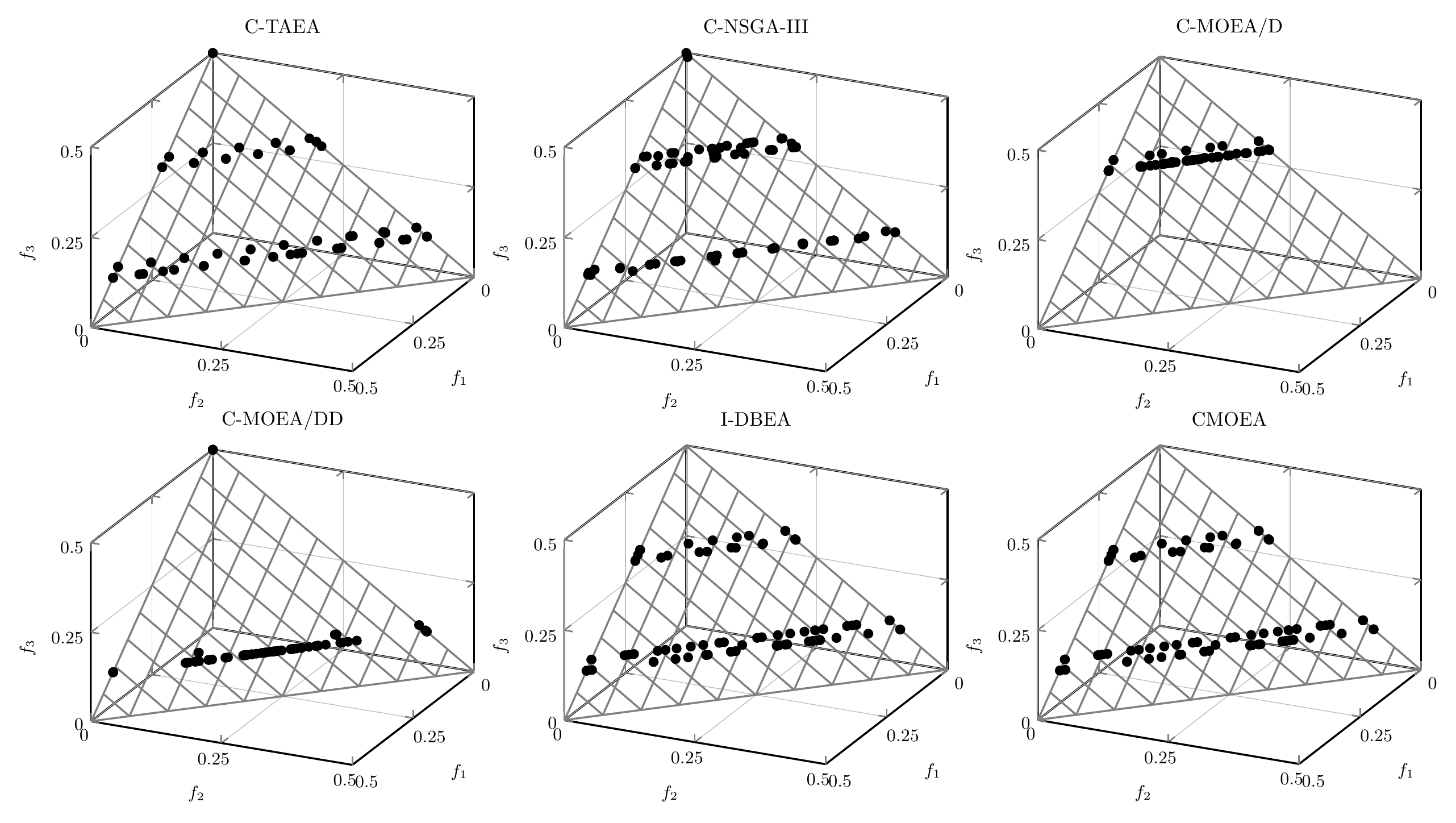}
    \caption{Scatter plots of the population obtained by C-TAEA and the peer algorithms on DC1-DTLZ1 (median IGD value).}
    \label{fig:DC1DTLZ1}
\end{figure}

\begin{figure}[htbp]
	\centering
    \includegraphics[width=.7\linewidth]{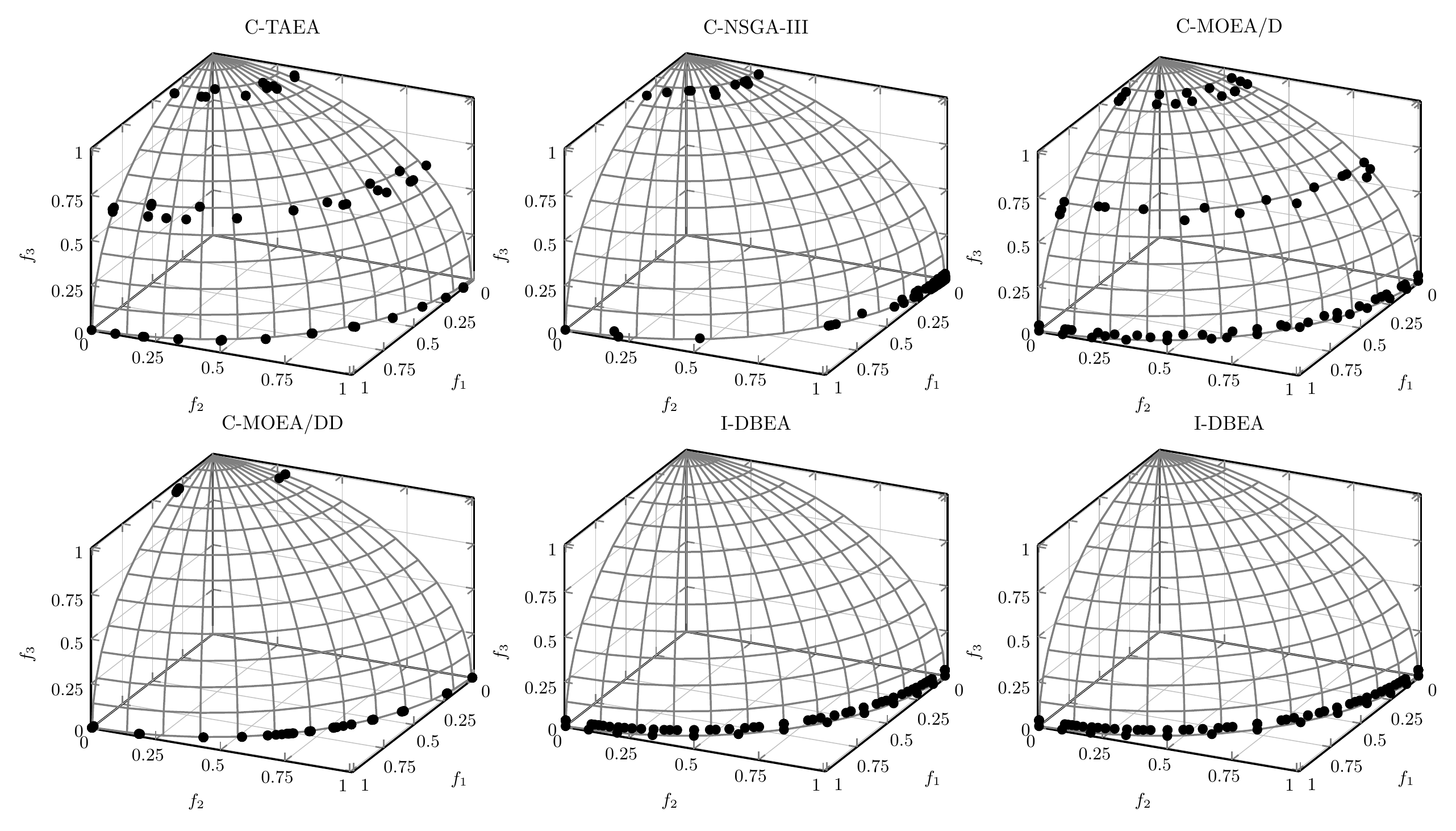}
    \caption{Scatter plots of the population obtained by C-TAEA and the peer algorithms on DC1-DTLZ3 (median IGD value).}
    \label{fig:DC1DTLZ3}
\end{figure}

Let us start from the Type-1 constrained problem. As described in Section I-B1) of the supplementary document, the constraints restrict the feasible region to a couple of narrow cone-shaped strips. Similar to C2-DTLZ2, the other peer algorithms have a risk of being trapped in one feasible region thus fail to find all feasible PF segments. However, DC1-DTLZ1 and DC1-DTLZ3 seem to be less challenging than C2-DTLZ2 with a small $r$ setting, given the observation that some peer algorithms are able to find a good number of solutions in different feasible PF segments as shown in~\pref{fig:DC1DTLZ1} and~\pref{fig:DC1DTLZ3}. This might be attributed to the $g(\mathbf{x})$ function of the baseline test problems, i.e., DTLZ1 and DTLZ3, which can make the crossover and mutation generate offspring far apart from their parents. Therefore, we can expect that solutions have some opportunities to jump over the locally feasible region. Nevertheless, as shown in~\pref{tab:DCDTLZ-IGD} and~\pref{tab:DCDTLZ-HV}, the IGD and HV values obtained by our proposed C-TAEA constantly outperform the other peer algorithms and the better results are with a statistical significance. 

The Type-2 constrained problem seems to be similar to C1-DTLZ1, at first glance, as shown in Fig. 8 and Fig. 9 of the supplementary document, where the constraints make the feasible region be reduced to a thin ribbon zone above the PF. However, it is more challenging due to the fluctuation in the CV of an infeasible solution when it approaches the PF, as shown in Fig. 10 of the supplementary document. As shown in~\pref{fig:DC2DTLZ1} and~\pref{fig:DC2DTLZ3}, we can clearly see that all other peer algorithms are trapped in a region far away from the PF. As the problem definitions of DC2-DTLZ1 and DC2-DTLZ3 shown in the supplementary document, all solutions obtained by the other peer algorithms are infeasible. Their failures on this type of constrained problems can be attributed to their feasibility-driven selection mechanisms, which drive the population fluctuate between the CV's local optima. As for our proposed C-TAEA, its success can be owed to the use of the DA. In particular, the selection mechanism of the DA does not take the CV into account so that it has sufficient selection pressure to move toward the PF. As shown in~\pref{fig:DC2DTLZ1} and~\pref{fig:DC2DTLZ3}, only C-TAEA finally find solutions on the PF.

\begin{figure}[htbp]
   	\centering
	\includegraphics[width=.7\linewidth]{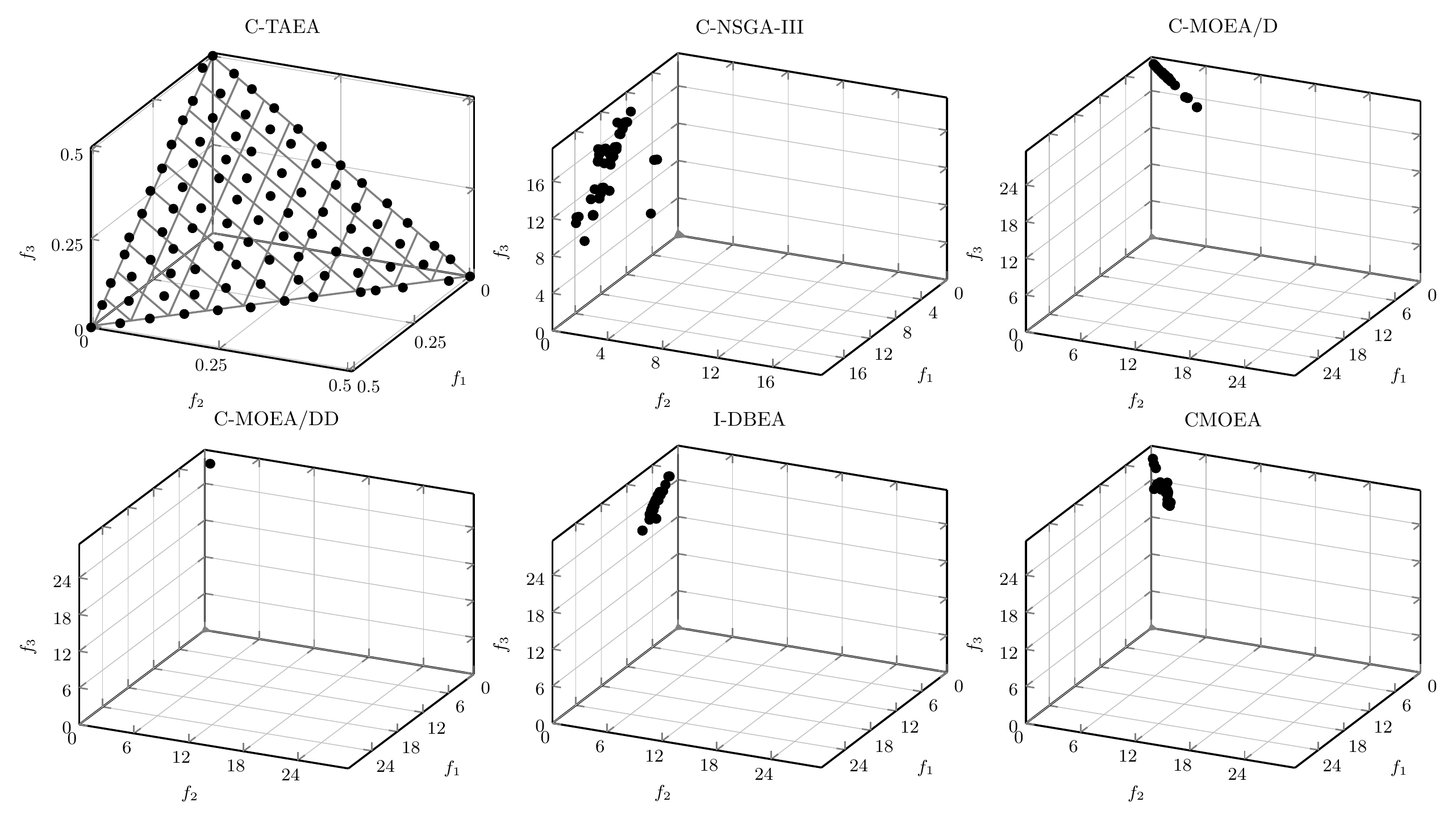}
    \caption{Scatter plots of the population obtained by C-TAEA and the peer algorithms on DC2-DTLZ1 (median IGD value).}
    \label{fig:DC2DTLZ1}
\end{figure}

\begin{figure}[htbp]
    \centering
    \includegraphics[width=.7\linewidth]{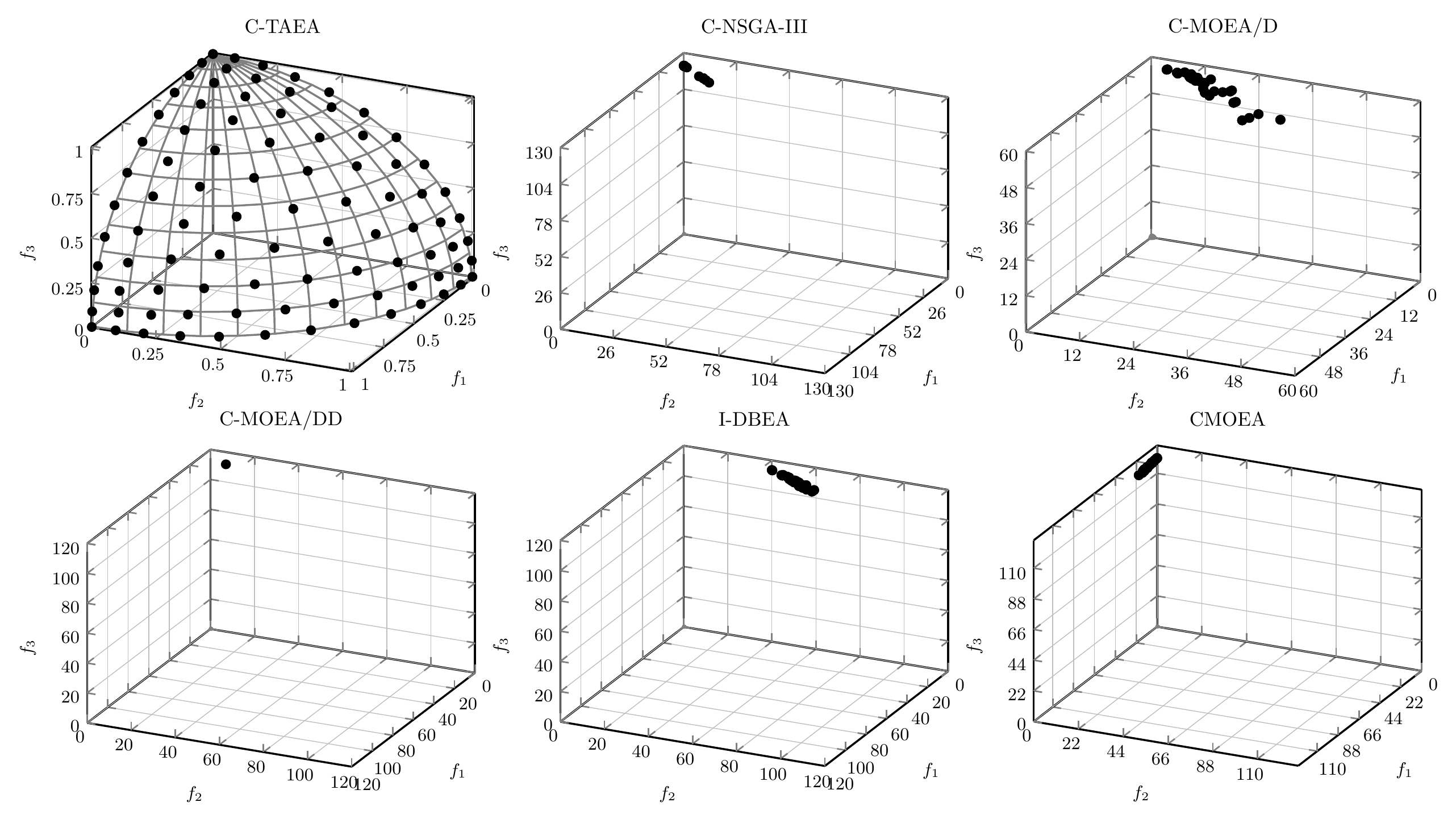}
    \caption{Scatter plots of the population obtained by C-TAEA and the peer algorithms on DC2-DTLZ3 (median IGD value).}
    \label{fig:DC2DTLZ3}
\end{figure}

\begin{figure}[htbp]
    \centering
    \includegraphics[width=.7\linewidth]{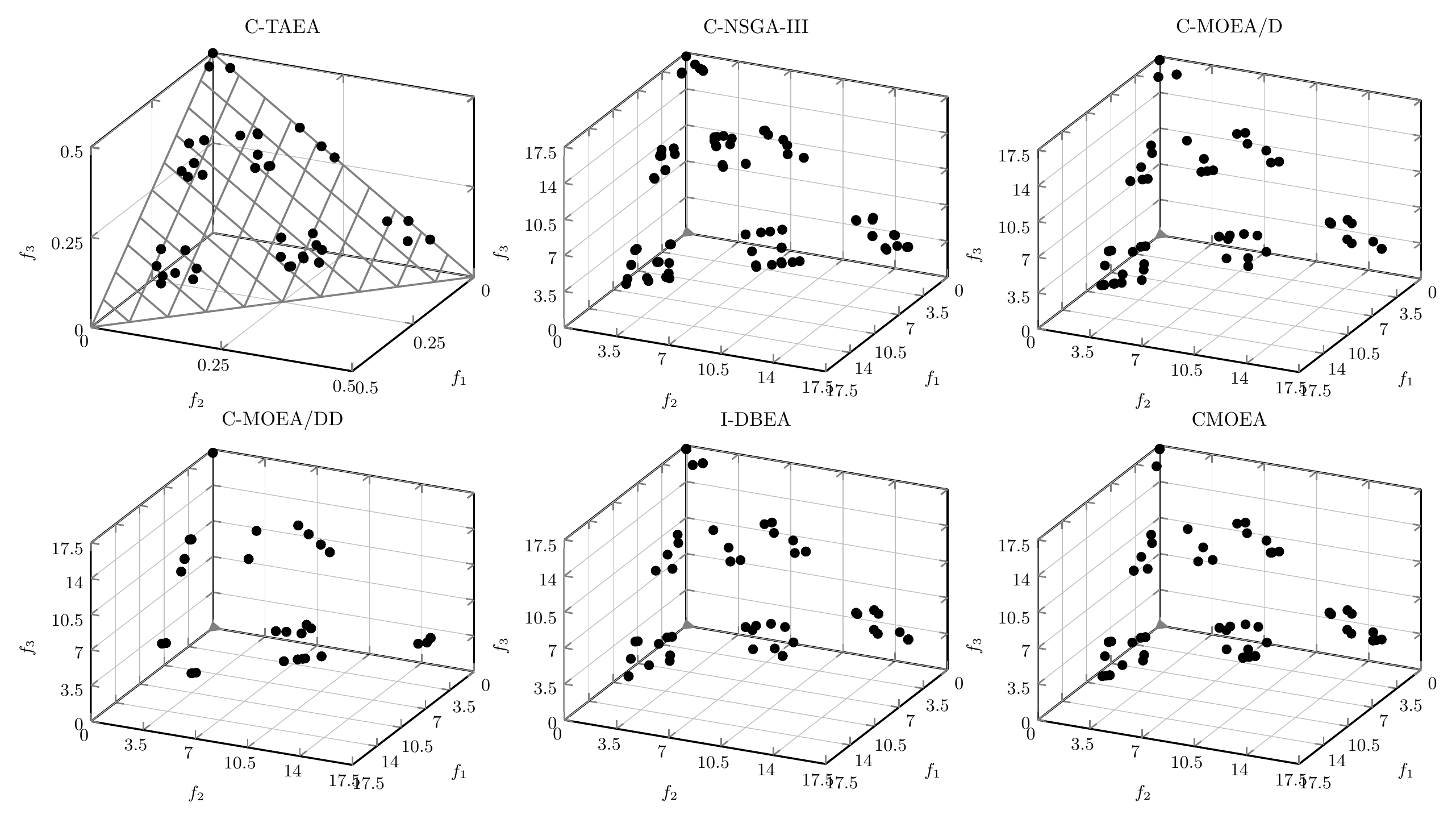}
    \caption{Scatter plots of the population obtained by C-TAEA and the peer algorithms on DC3-DTLZ1 (median IGD value).}
    \label{fig:DC3DTLZ1}
\end{figure}

\begin{figure}[htbp]
	\centering
    \includegraphics[width=.7\linewidth]{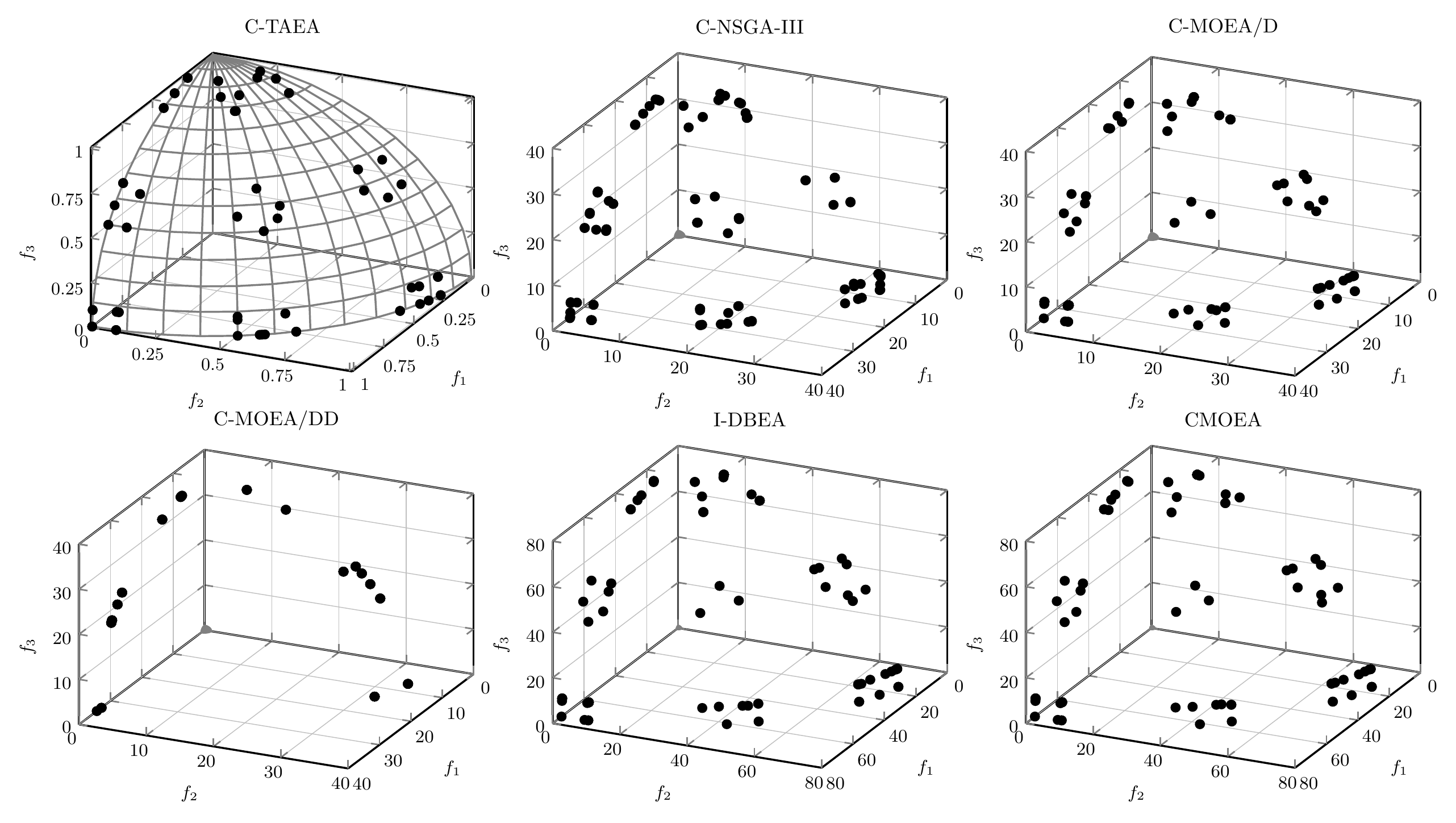}
    \caption{Scatter plots of the population obtained by C-TAEA and the peer algorithms on DC3-DTLZ3 (median IGD value).}
    \label{fig:DC3DTLZ3}
\end{figure}

As for the Type-3 constrained problem, its constraints are a combination of the previous two. In particular, the feasible regions are restricted to a couple of segmented cone stripes. In addition, there exists the same fluctuation, as the Type-2 constrained problem, in the CV of an infeasible solution when it approaches the PF. In this case, the other peer algorithms are not only struggling on jumping over a particular locally feasible region, but also have a significant trouble with the fluctuation (back and forth) of the population. Again, the success of our proposed C-TAEA is also attributed to the collaborative and complementary effects of two co-evolving archives. As shown in~\pref{fig:DC3DTLZ1} and~\pref{fig:DC3DTLZ3}, only C-TAEA finds all feasible PF segments while the other peer algorithms are stuck at some locally feasible regions away from the PF.

\subsection{Further Analysis}
\label{sec:analysis}

\begin{figure}[htbp]
\centering
\includegraphics[width=.7\linewidth]{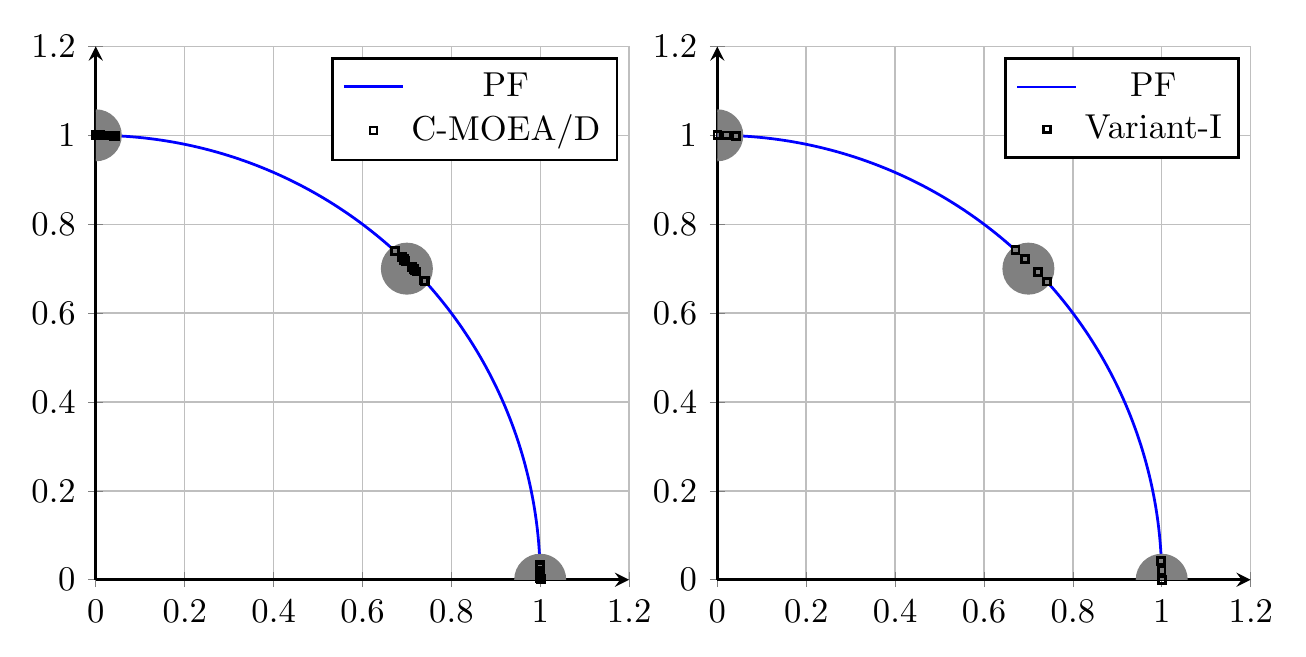}
\caption{Comparative results on the two-objective C2-DTLZ2.}
\label{fig:variant1}
\end{figure}

From the experimental results shown in~\pref{sec:CDTLZ} and~\pref{sec:DCDTLZ}, we have witnessed the superior performance of C-TAEA for solving various constrained multi-objective benchmark problems. To have a better understanding of its design principles, this subsection will investigate some important algorithmic components of C-TAEA by comparing it with the following two variants.
\begin{itemize}
\item\underline{Variant-I}: As shown in lines 15 to 21 of~\pref{alg:updateCA}, we iteratively remove the worst solution from the most crowded region when updating the CA. In particular, the worst solution is determined in terms of both its local crowdedness and its fitness value as defined in~\pref{eq:fitness}. This operation mainly aims to further improve the population diversity. To validate its effectiveness, we develop a variant in which the worst solution is simply defined as the one having the worst fitness value within the currently identified most crowded region.
\item\underline{Variant-II}: We claimed that the collaboration between the CA and DA is partially implemented by the restricted mating selection that automatically chooses the appropriate mating parents for offspring reproduction according to their evolution status. To validate the effectiveness of this operation, we develop another variant that randomly chooses mating parents from the CA and DA with an equal probability.
\end{itemize}
In the empirical studies, we use the same parameter settings as~\pref{sec:CDTLZ} and~\pref{sec:DCDTLZ} and compare the performance of C-TAEA with these two variants on C-DTLZ and DC-DTLZ benchmark problems. From the comparison results, i.e., the IGD and HV values respectively shown in Table IV and Table V of the supplementary document, we can see that the performance of C-TAEA and its two variants are comparable when the constraints are not difficult to solve, e.g., C1-DTLZ1, C3-DTLZ1/DTLZ4; whereas the superiority of C-TAEA becomes evident otherwise. More specifically, we find that Variant-I fails to maintain a good diversity when the feasible region is a small segment, e.g., C2-DTLZ2, DC1-DTLZ1/DTLZ3, DC3-DTLZ1/DTLZ3. \pref{fig:variant1} shows a comparison of the solutions found by C-TAEA and Variant-I on C2-DTLZ2 with $r=0.1$. From this figure, we can see that the solutions found by Variant-I are sparsely distributed within the feasible region. This is because the purely fitness-based selection strategy tends to drive solutions toward the corresponding weight vector within the feasible region as much as possible.

As for Variant-II, its random mating selection mechanism does not take enough advantage of the complementary effect of the CA and DA, thus it fails to help the algorithm overcome the locally infeasible barrier, e.g., C1-DTLZ3, DC2-DTLZ1/DTLZ3, DC3-DTLZ1/DTLZ3.

\subsection{Case Study: Water Distribution Network Optimization}
\label{sec:water}

Having tested C-TAEA's ability in solving various kinds of constrained benchmark problems, this section tends to investigate the performance of C-TAEA and the other peer algorithms on a real-world case study about optimal design of the water distribution network (WDN). In the past decade, multi-objective optimal design and rehabilitation of a WDN has attracted an increasing attention~\cite{Mala-JetmarovaS17}. The shift from the least-cost design to a multi-objective performance-based design advances decision makers' understanding of trade-off relationship between conflicting design objectives~\cite{Walski01}.

\begin{figure}[htbp]
    \centering
    \includegraphics[width=.7\linewidth]{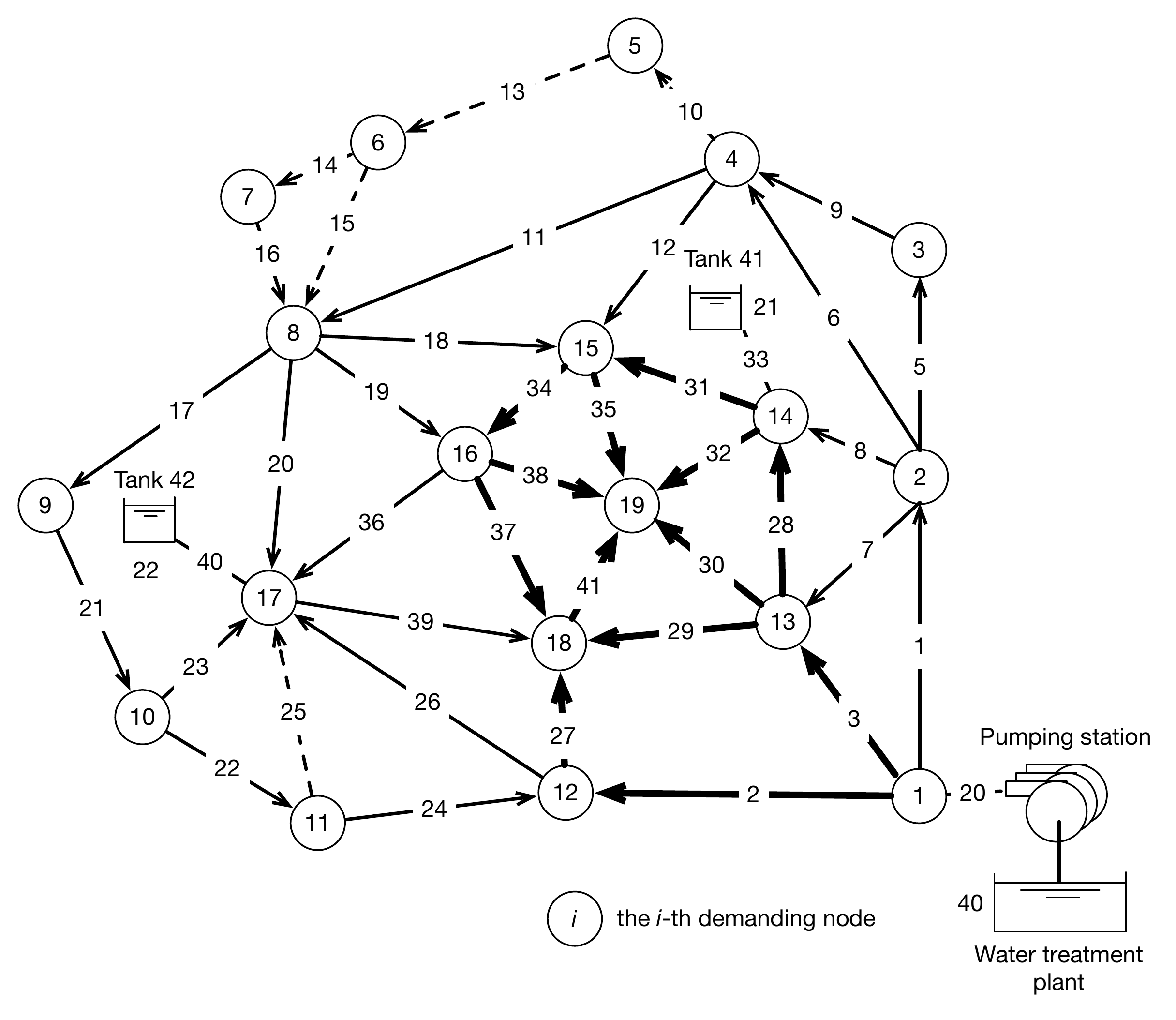}
    \caption{Layout of the anytown WDN.}
    \label{fig:anytown}
\end{figure}

This paper uses the Anytown WDN, one of the most popular benchmark networks, as the case study. Anytown WDN has many typical features and challenges that can be found in real-world networks, e.g., pump scheduling, tank storage provision, and fire-fighting capacity provision. The network layout is shown in~\pref{fig:anytown}, where it has 35 pipes, 2 storage tanks, and 3 identical pumps delivering water from the treatment plant into the system. To meet the city expansion and increasing demands, 77 decision variables are considered, including 35 variables related to the existing pipes (with options of cleaning and lining or duplication with a parallel pipe), six new pipe diameters, 12 variables for two potential tanks, and 24 variables for the number of pumps in operation during 24 hours of a day. In this paper, the WDN design problem is formulated as a four-objective optimization problem with two constraints. In particular, we consider costs, resilience index, statistical flow entropy and water age as the objective functions. More detailed descriptions of the problem formulation can be found in Section IV of the supplementary document.

\begin{figure}[htbp]
	\centering
    \includegraphics[width=.7\linewidth]{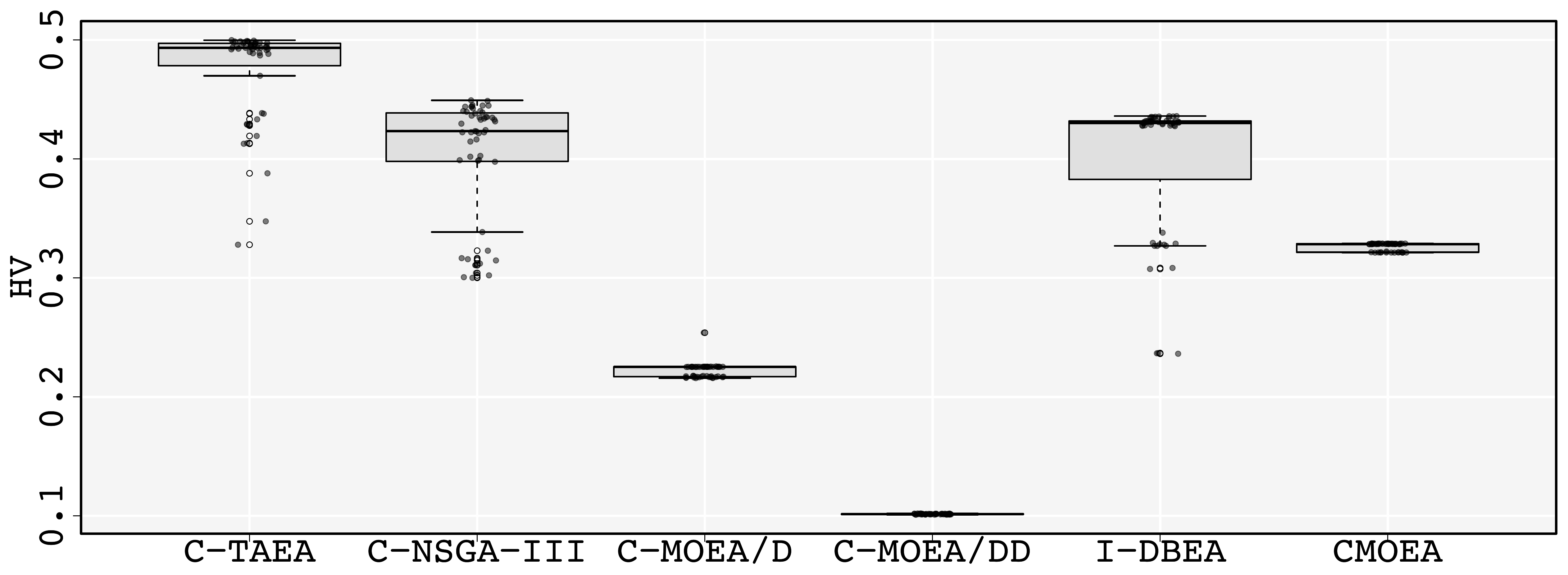}
    \caption{Box plots of HV obtained by different algorithms.}
    \label{fig:boxplot}
\end{figure}

In the experiment, C-TAEA and the other five peer algorithms use the solution encoding scheme as suggested in~\cite{FuKKR13}. The population size is set to $N=100$, and the number of function evaluations used for each algorithm is set to $10,000\times N$. The reproduction operators and their corresponding parameters are still set the same as before. Since the true PF is unknown for this real-world WDN model, we only use the HV as the performance metric where $\mathbf{z}^r=(1.1,\cdots,1.1)^T$. In particular, we normalize the objective functions before calculating the HV metric. From the box plots (with respect to 51 independent runs) shown in~\pref{fig:boxplot}, we can clearly see that our proposed C-TAEA shows better performance than the other five peer algorithms.

%% file: conclusion.tex

\section{Conclusions and Future Directions}
\label{sec:conclusions}

In this paper, we have suggested a parameter-free constraint handling technique, a two-archive evolutionary algorithm (C-TAEA), for constrained multi-objective optimization. In C-TAEA, we simultaneously maintain two co-evolving archives. Specifically, one population, denoted as CA, mainly focuses on driving the population toward the PF; while the other one, denoted as DA, mainly tends to explore the areas under-exploited by the CA (even those infeasible regions) thus provide more diversified information. In this case, the CA and DA have different behaviors and complementary effects. In particular, they complement each other via a restricted mating selection mechanism which selects complementary mating parents for offspring reproduction. The performance of C-TAEA has been investigated on a series of benchmark problems with various types of constraints and up to 15 objectives. The empirical results fully demonstrate its competitiveness on CMOPs, in comparison to five state-of-the-art constrained EMO algorithms. In addition to artificial benchmark problems, the effectiveness of C-TAEA has also been validated on a real-world case study of the WDN design optimization. 

As previously also demonstrated in~\cite{PraditwongY06,LiLTY14,WangJY15}, we believe that C-TAEA is more than a specific algorithm. Instead, its basic idea, co-evolving multiple complementary archives, can be widely used in the general EMO algorithm design. In future, it is worth further investigating its underlying mechanisms from both algorithm design and theoretical foundation perspectives. Furthermore, we plan to investigate the effectiveness of this two-archive co-evolving framework on a wider range of problems, such as unconstrained MOP including those with complex properties (e.g., problems with complecated PSs~\cite{UF} and imbalanced convergence and diversity~\cite{LiuCDG17}), dynamic optimization (e.g., problems with a changing number of objectives or constraints~\cite{ChenLY17}), and other real-world applications.